\documentclass[11pt]{article}

\usepackage[]{acl}
\usepackage[T1]{fontenc}
\usepackage{times}
\usepackage{latexsym}

\usepackage[utf8]{inputenc}
\usepackage{microtype}
\usepackage{booktabs}
\usepackage{tablefootnote}
\usepackage{xcolor}
\usepackage{multirow}
\usepackage{hyperref}
\usepackage{xspace}
\usepackage{caption}
\usepackage{subcaption}
\usepackage{amsmath}
\usepackage{float}
\usepackage{colortbl}
\usepackage{enumitem}
\usepackage{adjustbox}
\usepackage{pifont}
\usepackage{tabularx}

\def\dataset/{V\=arta}
\def\rouge/{\textsc{rouge}}
\def\rougeone/{\textsc{rouge-1}}
\def\rougetwo/{\textsc{rouge-2}}
\def\rougel/{\textsc{rouge-l}}

\newcommand{\cmark}{\ding{51}}
\newcommand{\xmark}{\ding{55}}

\title{\dataset/: A Large-Scale Headline-Generation Dataset for Indic Languages}

\makeatletter
\newcommand{\printfnsymbol}[1]{%
  \textsuperscript{\@fnsymbol{#1}}%
}
\makeatother

\author{Rahul Aralikatte$^{1, 2}$\thanks{~~Equal contribution. Corresponding author: Rahul Aralikatte (\href{mailto:rahul.aralikatte@mila.quebec}{rahul.aralikatte@mila.quebec})}~~~~Ziling Cheng$^{1,2}$\printfnsymbol{1} \\ \textbf{Sumanth Doddapaneni}$^{3,4}$~~~~\textbf{Jackie Chi Kit Cheung}$^{1, 2, 5}$ \\   
    $^1$Mila -- Quebec Artificial Intelligence Institute~~~$^2$McGill University\\
    $^3$IIT Madras~~~$^4$AI4Bharat~~~$^5$Canada CIFAR AI Chair \\
}

\begin{document}
\maketitle
\begin{abstract}
We present \dataset/, a large-scale multilingual dataset for headline generation in Indic languages. This dataset includes 41.8 million news articles in 14 different Indic languages (and English), which come from a variety of high-quality sources. To the best of our knowledge, this is the largest collection of curated articles for Indic languages currently available. We use the data collected in a series of experiments to answer important questions related to Indic NLP and multilinguality research in general. We show that the dataset is challenging even for state-of-the-art abstractive models and that they perform only slightly better than extractive baselines. Owing to its size, we also show that the dataset can be used to pretrain strong language models that outperform competitive baselines in both NLU and NLG benchmarks. The data and models are available at \url{https://github.com/rahular/varta}.
\end{abstract}

\section{Introduction}\label{sec:intro}
Headline generation is a special case of abstractive summarization where the goal is to create a brief, often single-sentence `summary' of a news article. Unlike traditional summaries, which are a few sentences long and concisely convey the most important information from an article, a headline typically highlights one key fact from the article that is considered to be the most significant. In recent years, headlines have also been written to be catchy and increase click-through rates. This trend has made the task harder, as the lexical overlap between headlines and articles is low, while their compression ratio is high.

There are several datasets available to train and evaluate headline generation and abstractive summarization models, but they are largely limited to English \citep{xsum, cnndm, gigaword, tldr}. Although there have been efforts to create multilingual datasets such as MLSum \citep{mlsum} and XLSum \citep{xlsum}, the representation of Indic languages in these datasets is still minimal.\footnote{The only exception to this is the recently released headline-generation dataset proposed by \citet{indichl}. We will compare this with our dataset in \S\ref{sec:related-work}.} As a result, it is challenging to create good headline generation or summarization systems for these languages, which are spoken by approximately one in five people worldwide.

\begin{table}
\setlength{\tabcolsep}{4pt}
\small
\begin{tabular}{rcccc}
    \toprule
     & \multirow{ 2}{*}{MLSum} & \multirow{ 2}{*}{XLSum} & Indic- & \multirow{ 2}{*}{\dataset/} \\
     & & & Headline & \\
     \midrule
    \# of langs. & 5 & 44 & 11 & 15 \\
    \# of Indic langs. & 0 & 9 & 11 & 14 \\
    Headline present & \xmark & \cmark & \cmark & \cmark \\
    Summary present & \cmark & \cmark & \xmark & \xmark \\
    Size of Indic parts & 0 & 165K & 1.3M & 34.5M \\
    Total size & 1.5M & 1M & 1.3M & 41.8M \\
    \bottomrule
\end{tabular}
\caption{Comparison of existing multilingual headline generation and summarization datasets with \dataset/.}
\label{tab:intro}
\end{table}

\paragraph{Motivation} Indic headline generation is particularly interesting because this family of languages presents unique challenges. For example, though most Indic languages are closely related, they use different scripts which makes transfer learning harder. Many of the languages are morphologically rich, making headlines compact and thus harder to predict. With few exceptions, most Indic languages are low-resource and are underrepresented in multilingual models, which makes few-shot learning ineffective. Recent efforts have shown that language family-specific pretraining enables better transfer among languages \cite{indicxtreme} and that quality data is required in large quantities for good downstream performance on text generation tasks like summarization \cite[][HugeNews]{pegasus}. 

Thus, there is a strong need for a multilingual, large-scale, high-quality dataset for Indic languages. To fill this gap, we introduce \dataset/, a dataset consisting of more than 41 million article-headline pairs in 15 languages (14 Indic languages + English). The data is crawled from DailyHunt, a popular news aggregator in India that pulls high-quality articles from multiple trusted and reputed news publishers. It also covers a wide range of topics such as politics, science, entertainment, sports, business, etc., which makes the dataset diverse both in terms of domains and writing styles. Table \ref{tab:intro} compares \dataset/ with other multilingual summarization and headline generation datasets.

\paragraph{Contributions} The main contribution of this work is the \dataset/ dataset, which we hope will push the research on headline generation and summarization in Indic languages. We run two sets of experiments: (i) The first set of experiments deals only with the task of headline generation under different training and evaluation settings. By analyzing the results of these experiments, we answer several important research questions related to text generation from an Indic perspective. (ii) Owing to the size and multilingual nature of \dataset/, the second set of experiments treats the dataset as a pretraining corpus. We pretrain a BERT-style encoder-only model and a T5-style encoder-decoder model and show that the models outperform strong baselines on IndicXTREME \cite{indicxtreme} and IndicNLG \cite{indicnlg} benchmarks, respectively. Finally, we release the data and the pretrained models so that the community can build on top of our research.

\section{Related Work}\label{sec:related-work}
\paragraph{English Data} There have been several proposed datasets for the generation of English headlines and other similar tasks. The DUC 2003 and 2004 datasets, which contain 500 and 624 article-headline pairs from the Associated Press and New York Times, respectively, were among the first to be released. \citet{gigaword} introduced a dataset based on the Gigaword corpus \cite{napoles2012annotated}, which pairs the first sentence of a news article with its headline for sentence summarization. This dataset has around 4 million pairs in total. Other notable datasets from the literature include XSUM \citep{xsum} for 1-2 sentence summaries of BBC articles, the Google dataset for sentence compression \citep{filippova-altun-2013-overcoming}, and the TLDR corpus \citep{tldr} for summarizing Reddit posts.

\paragraph{Multilingual Data} The Columbia Newsblaster \citep{evans2004columbia} was one of the first datasets to include multilingual data for summarization. It includes around 500 news articles in English, Russian, and Japanese, with annotations for articles, texts, titles, images, and captions. In recent years, larger datasets with a greater variety of languages have been curated. MLSum \citep{mlsum} comprises 1.5 million pairs of articles and summaries in French, German, Spanish, Russian, and Turkish. XLSum \citep{xlsum} follows the format of XSUM and crawls the BBC websites to obtain 1 million pairs in 44 languages.\footnote{The BBC publishes articles in 44 languages, nine of which are Indic.} This dataset also includes the headlines of the articles, along with their summaries. \citet{indichl} propose a headline generation dataset specifically for Indic languages, which includes 1.3 million data points in 11 Indic languages.

\paragraph{Modeling Approaches} The task of headline generation has been approached in several ways over the years. An early approach was proposed by \citet{Banko2000HeadlineGB}, who viewed the task as a machine translation problem. Subsequently, as sequence-to-sequence models became more popular \cite{gigaword}, many works have used encoder-decoder architectures to tackle this problem \citep{shen2017recent}. More recently, pretrained language models such as BART \citep{bart} and T5 \cite{t5} have been used for headline generation and other related tasks with great success.
    
\section{Data}
\dataset/ contains 41.8 million high-quality news articles in 14 Indic languages and English. With 34.5 million non-English article-headline pairs, it is the largest headline-generation dataset of its kind. In this section, we detail how the dataset is collected along with the various processing steps involved, before moving on to some interesting statistics and analysis of the data. Table \ref{tbl:app-data-example} in the Appendix showcases some randomly selected articles from \dataset/.

\subsection{DailyHunt}\label{sec:dailyhunt}
\paragraph{Source}
DailyHunt is a popular aggregator platform for news in India.\footnote{\url{https://m.dailyhunt.in/}} It curates articles from over 1773 publishers in English and 14 Indic languages: Assamese, Bhojpuri, Bengali, Gujarati, Hindi, Kannada, Malayalam, Marathi, Nepali, Oriya, Punjabi, Tamil, Telugu, and Urdu. We collect all our data from this platform and restrict ourselves to articles that were published between January 2010 and August 2022.

\paragraph{Data Collection} Since DailyHunt does not have an external facing API, we crawl their website using Scrapy,\footnote{\url{https://scrapy.org/}} a Python-based scraping tool that collects data efficiently without burdening their servers. To maintain the quality of the collected data, we: (i) discard articles that are less than 50 words long, (ii) discard articles where an image or video is prominently placed (since these articles cannot be understood without the aid of the embedded media), and (iii) discard articles which require us to navigate to the publisher website to get the entire text.

\begin{table*}
\small
\centering
\setlength{\tabcolsep}{3pt} %
\begin{tabular}{lcccccccccccccc} 
\toprule
\multirow{2}{*}{Lang.} & \multirow{2}{*}{Size} &\multirow{2}{*}{Domain} &\multirow{2}{*}{Vocab} & \multicolumn{4}{c}{Ratio of novel n-grams (\%)} & \multirow{2}{*}{CR } &  \multicolumn{2}{c}{$|$article$|$} &  \multicolumn{2}{c}{$|$headline$|$} & \multirow{2}{*}{Lead-1} & \multirow{2}{*}{Ext Or.} \\ 
\cmidrule(lr){5-8} \cmidrule(lr){10-11} \cmidrule(lr){12-13}
 && Count & Size& 1-gram & 2-gram & 3-gram & 4-gram & (\%) & Tok. & Sent. & Tok. & Sent. & & \\
\midrule
as & 87.5K & 20 & 964K & 33.59 & 60.67 & 72.76 & 78.16 & 8.01 & 207.79 & 12.05 & 11.76 & 1.1 & 18.41 & 35.57 \\
bh & 1.56K & 2 & 62.1K & 23.61 & 58.56 & 74.08 & 80.03 & 5.37 & 446.08 & 23.07 & 14.31 & 1.03 & 27.11 & 35.26 \\
bn & 2.25M & 155& 6.00M & 33.65 & 67.55 & 80.71 & 85.58 & 5.22 & 290.9 & 21.58 & 12.15 & 1.19 & 20.2 & 35.62  \\
en  & 7.27M & 488 & 17.9M & 28.98 & 65.23 & 81.18 & 87.43 & 4.15 & 467.94 & 17.92 & 14.21 & 1.05 & 23.78 & 32.83 \\
gu & 2.00M & 105 & 7.49M & 33.69 & 67.9 & 80.78 & 86.18 & 5.52 & 325.94 & 17.74 & 13.83 & 1.1 & 14.59 & 30.55 \\
hi & 14.4M & 413 & 19.1M & 19.58 & 57.84 & 76.06 & 84.95 & 5.06 & 375.93 & 17.59 & 15.51 & 1.03 & 17.71 & 35.85 \\
kn & 1.47M & 74 & 8.44M & 34.83 & 68.43 & 82.63 & 87.35 & 6.2 & 233.41 & 16.97 & 10.29 & 1.12 & 19.36 & 32.4 \\
ml & 3.47M & 136 & 20.6M & 30.62 & 59.3 & 72.12 & 76.14 & 7.28 & 178.02 & 14.8 & 10.04 & 1.07 & 32.41 & 42.07 \\
mr & 2.67M & 159 & 10.1M & 33.8 & 67.99 & 81.51 & 86.44 & 4.6 & 321.77 & 20.77 & 11.94 & 1.15 & 14.53 & 31.33 \\
ne & 32.5K & 2 & 435K & 26.37 & 62.97 & 79.89 & 88.37 & 4.99 & 253.17 & 16.06 & 9.45 & 1.01 & 4.25 & 36.38 \\
or & 1.09M & 59 & 3.43M &28.75 & 64.3 & 78.96 & 84.07 & 6.12 & 214.66 & 14.49 & 10.26 & 1.06 & 21.52 & 35.89 \\
pa & 842K & 32 & 2.10M & 20.16 & 57.1 & 73.95 & 82.42 & 5.81 & 328.87 & 12.26 & 14.62 & 1.03 & 22.24 & 31.75 \\
ta & 2.64M & 120 & 10.0M & 36.21 & 67.63 & 81.58 & 86.59 & 7.18 & 210.73 & 14.94 & 11.43 & 1.64 & 23.77 & 34.23 \\
te & 3.27M & 113 & 13.4M & 37.94 & 71.22 & 82.2 & 79.8 & 5.13 & 238.62 & 19.04 & 9.17 & 1.25 & 16.67 & 30.31 \\
ur & 303K & 21 & 1.56M & 15.2 & 47.67 & 66.08 & 76.23 & 5.08 & 442.61 & 15.51 & 14.29 & 1.08 & 25.91 & 35.08 \\
\midrule
Avg. & - & - & - & 29.13 & 62.96 & 77.63 & 83.32 & 5.72 & 302.43 & 16.99 & 12.22 & 1.13 & 20.16 & 34.34 \\
\bottomrule
\end{tabular}
\caption{Statistics and extractive baseline results on \dataset/. CR and Ext Or. stand for the Compression Ratio and Extractive Oracle, respectively. Only \rougel/ F1-scores are reported for Lead-1 and Extractive Oracle; see Table \ref{tbl:app-baseline} in the Appendix for \rougeone/ and \rougetwo/ scores.}
\label{tbl:stats}
\end{table*}

\paragraph{Processing} We only retain text from the articles and remove other HTML content like embedded media, and hyperlinks to other websites. We also de-duplicate articles by using the headline as the identifier. We do not use the URLs as identifiers since DailyHunt re-aggregates an article if the original article text is changed by the publisher. We find and remove around 7.2 million such `stale' articles, and keep only the latest versions.\footnote{These duplicates themselves form an interesting dataset of how news articles are edited over time.}

\subsection{Statistics}
The language-wise sizes of the processed data are shown in Table~\ref{tbl:stats}. To better understand the various properties of the dataset, we also compute the following metrics for each language: (i) the ratio of novel n-grams between the headline and the article, (ii) the average article and headline lengths, and (iii) the compression ratio between the headline and the article. Additionally, we also report the number of distinct words (`Vocab Size' in Table \ref{tbl:stats}), and the number of publishers (`Domain Count' in Table \ref{tbl:stats}) for each language.

\paragraph{Novel n-grams}
The ratio of novel n-grams between a headline and its article text is a proxy for the abstractiveness of the headline, with higher values indicating more abstract headlines. Across languages, 29\% and 63\% of the unigrams and bigrams are unique respectively. In 7 of the 15 languages, the novel unigram ratio exceeds $1/3$ indicating a high degree of abstraction and low lexical overlap. 

\paragraph{Article and Headline lengths}
On average, \dataset/ articles have 17 sentences, and the headlines have just over one sentence. A typical article sentence contains about 18 words, and a headline sentence contains 11 words. This indicates that the headlines in our dataset are usually 39\% smaller than the average sentence in an article. This gives us an idea about the extreme nature of summarization in the dataset.

\paragraph{Compression Ratio}
The ratio between the number of tokens in the headline and the article gives a sense of the conciseness of the headline, with lower values indicating more concise headlines. Overall, \dataset/ has a compression ratio of 5.72\%, with English and Assamese headlines being the most and least concise respectively. 

\subsection{Extractive Baselines}
We evaluate two extractive summarization methods on the data to see how far we can get by only selecting a sentence from the article as the headline: (i) Lead-1: choose the first sentence of the article as the headline. This can be seen as a strong performance lower bound \cite{nenkova-2005}, and (ii) Extractive Oracle: choose a sentence from the article which gives the highest \rouge/ score with respect to the gold headline. This provides an upper bound for extractive models \cite{10.5555/3298483.3298681}. We report the language-wise \rougel/ scores in Table~\ref{tbl:stats} with more detailed numbers in Table \ref{tbl:app-baseline} of the Appendix.

\subsection{Data Splits}\label{sec:data-splits}
From every language, we randomly sample 10,000 articles each for validation and testing. We also ensure that at least 80\% of a language's data is available for training. Therefore, if a language has less than 100,000 articles, we restrict its validation and test splits to 10\% of its size. Table \ref{tbl:app-lang-split} in the Appendix shows the splits for each language.

Since this training set has more than 41 million articles, performing experiments and ablations will be compute- and time-intensive. Therefore, we create a \textsc{small} training set by limiting the number of articles from each language to 100K. This \textsc{small} training set with a size of 1.3M is used in all our fine-tuning experiments.

\section{Experiments}
Rather than trying to squeeze optimal performance from models, the aim of our experiments is to perform a detailed study of how different training settings affect model behavior. In this section, we first present a list of research questions that our experiments try to answer. We then provide a brief overview of the models used in the experiments. Lastly, we describe the different training and evaluation configurations of our experiments before presenting our findings in \S\ref{sec:results}. 

\subsection{Research Questions}\label{sec:rq}
\subsubsection*{RQ1: What is the best bridge language? }
A bridge or pivot language is an intermediary for transferring knowledge learned by training in one language to others. English is commonly used as the bridge language in literature. We test if English is indeed the best bridge language in our data, or if can we use Hindi since it is typologically similar to other languages in \dataset/. 
\subsubsection*{RQ2: Are different scripts a hindrance? }
\dataset/ contains languages with 11 unique scripts. Can we have a better transfer if we use a single script during fine-tuning? Which script would help the most? Why?
\subsubsection*{RQ3: Which setup performs the best?}
We can finetune models under different settings with different data configurations. Which of them would give us the best result? What is required for effective transfer? Does training language-family-specific models give us an advantage over massive multilinguality? 
\subsubsection*{RQ4: Can \dataset/ be used for pretraining?}
The objective of this work is to curate a high-quality dataset for headline generation in Indic languages. Upon realization that the final dataset size is comparable to existing pretraining corpora for Indic languages, we decided to compare the downstream performance of models pretrained on \dataset/ against those pretrained on similar corpora. Therefore, we ask the following questions: Can \dataset/ be used to pretrain both NLU and NLG models? If yes, are they competitive? Can our finetuned models generalize to other similar tasks, like abstractive summarization?
 
\subsection{Models}\label{sec:models}

\paragraph{mT5} introduced by \citet{xue-etal-2021-mt5}, it is the multilingual variant of T5 \cite{t5}. It is pretrained on the mC4 corpus, comprising 101 languages. The pretraining objective of mT5 is ``span-corruption'' which is generating spans of text that are masked in the input.

\paragraph{mBERT-seq2seq} uses a multilingual variant of BERT \cite{devlin-etal-2019-bert} pretrained on Wikipedia, on a total of 104 languages.\footnote{It should be noted that both mT5 and mBERT are not pretrained on 3 \dataset/ languages: \textit{as}, \textit{bh}, and \textit{or}.} Since BERT is an encoder-only model trained on masked language modeling, it cannot be used directly for generation. However, \citet{rothe-etal-2020-leveraging} show that initializing encoder-decoder models with pretrained encoder-only or decoder-only checkpoints or ``warm-starting'', yields competitive results in multiple sequence-to-sequence tasks. We warm-start both the encoder and decoder of our model with mBERT weights and the encoder-decoder attention weights are initialized randomly.\footnote{We do not share parameters between the encoder and decoder since early experiments showed that sharing parameters slightly hurt performance.} We denote this model as just mBERT from here on.

Another reason for choosing such a model configuration as a baseline is that we can compare and contrast how models learn when their encoder-decoder attention weights are (not) initialized from pretrained weights. \citet{aralikatte-etal-2021-focus} show models with randomly initialized encoder-decoder attention learn data characteristics much better than their pre-initialized counterparts. 

\paragraph{\dataset/-T5} In the pretraining corpus used for mT5 and mBERT, the size of the Indic languages we are interested in, is relatively small. Previous works have shown that such underrepresented languages in massively multilingual models suffer due to a lack of model capacity \cite{evaluating-inclusivity}, and poor transfer \cite[][Section 2]{ponti-etal-2021-minimax}, and that pretraining only on a group of closely related languages results in better downstream performance \cite{indicxtreme}. Therefore, we use the full training set from \dataset/ to pretrain a T5 model from scratch.

We use span corruption and gap-sentence generation as the pretraining objectives. Both objectives are sampled uniformly during pretraining. Span corruption is similar to masked language modeling except that instead of masking random tokens, we mask spans of tokens with an average length of 3. In gap-sentence prediction, whole sentences are masked instead of spans. We follow the original work \citep{pegasus}, and select sentences based on their `importance'. \rougeone/ F1-score between the sentence and the document is used as a proxy for importance. We use 0.15 and 0.2 as the masking ratios for span corruption and gap-sentence generation, respectively.

We use a standard {\it T5-base} architecture with 12 encoder and decoder layers, and 12 attention heads. Since data sizes across languages in \dataset/ vary from 1.5K (Bhojpuri) to 14.4M articles (Hindi), we use standard temperature-based sampling to upsample data when necessary.\footnote{We use $\alpha=0.7$.} For other training details, see Appendix \ref{sec:pt-t5}.

\subsection{Finetuning and Evaluation settings}
To answer the questions posed in \S\ref{sec:rq}, we perform a series of experiments. First, we finetune each model described in \S\ref{sec:models} on \dataset/ in five settings:
(i) \textit{en}: finetune only on English data from the \textsc{small} training set, and evaluate on all language test sets in original scripts.
(ii) \textit{hi}: finetune only on Hindi data from the \textsc{small} training set, and evaluate on all language test sets in original scripts.
(iii) \textit{latin}: finetune on the \textsc{small} training set transliterated to Latin (English) script, and evaluate on all language test sets transliterated to Latin script.\footnote{We use the IndicTrans \cite{Bhat:2014:ISS:2824864.2824872} and IndicNLP \cite{kunchukuttan2020indicnlp} libraries to transliterate text to English and Devanagari respectively. For Urdu, we use IndicTrans in both cases. Though Bhojpuri uses Devanagari, we ignore it while performing the unified script experiments since it does not have a good transliteration system.}
(iv) \textit{dvn.}: finetune on the \textsc{small} training set transliterated to Devanagari script, and evaluate on all language test sets transliterated to Devanagari script.\footnote{We denote \textit{latin} (\textit{dvn.}) model as English (Devanagari) `unified model' since it is trained on a single, unified script.}
(v) \textit{all}: finetune on all languages of the \textsc{small} training set, and evaluate on all language test sets, in their original scripts.

Next, to assess the generalizability of our pretrained model, we conduct the following two experiments on the \dataset/-T5 model:
(i) evalaute on the XL-Sum dataset \citep{xlsum}. XL-Sum is particularly useful as it contains article-headline-summary triplets, which allows for the evaluation of both headline generation and abstractive summarization tasks. We evaluate the best models from the previous set of experiments in a zero-shot setting.
(ii) To determine the generalizability of \dataset/-T5 on text generation tasks, we evaluate the model on the IndicNLG benchmark \cite{indicnlg} which contains five diverse generation tasks in 11 Indic languages.

Finally, to determine if \dataset/ can be used to train good NLU models, we train \dataset/-BERT, an encoder-only model trained with the masked language modeling objective and evaluated on IndicXTREME \cite{indicxtreme}, a zero-shot cross-lingual NLU benchmark for Indic languages consisting of nine tasks in 18 languages.

\begin{table*}
\small
\centering
\setlength{\tabcolsep}{3pt} %
\begin{tabular}{c|r|ccccccccccccccc|c}
\toprule
\multicolumn{2}{c|}{Model} & as & bh & bn & en & gu & hi  & kn & ml & mr & ne & or & pa & ta & te & ur & Avg. \\
\midrule 
\multirow{5}{*}{\rotatebox[origin=c]{90}{mBERT}} & en & 1.24 & 2.82 & 1.92 & 39.14 & 1.70 & 4.44 & 2.73 & 1.75 & 2.88 & 2.29 & 0.35 & 2.44 & 2.04 & 3.42 & 2.75 & 4.79 \\
& hi & 1.41 & 29.86 & 1.35 & 15.50 & 1.35 & 39.89 & 1.35 & 1.01 & 19.35 & 29.74 & 0.26 & 1.89 & 1.07 & 2.36 & 2.17 & 9.90 \\
& latin & 28.04 & - & 24.92 & 34.89 & 21.15 & 28.98 & 28.10 & 35.52 & 21.53 & 31.14 & 28.08 & 32.52 & 27.36 & 24.20 & 35.72 & 28.72 \\
& dvn. & 30.66 & - & 28.24 & 39.37 & 24.88 & 34.55 & 32.09 & 37.13 & 26.91 & 38.80 & 33.29 & 37.48 & 30.77 & 27.40 & 40.62 & 33.01 \\
& all & 32.98 & 37.20 & 33.68 & 41.36 & 26.81 & 36.89 & 35.50 & 39.88 & 29.29 & 41.55 & 1.24 & 40.60 & 36.25 & 30.21 & 43.82 & 33.82 \\
\midrule
\multirow{5}{*}{\rotatebox[origin=c]{90}{mT5}} & en & 24.19 & 29.01 & 27.73 & 43.20 & 21.66 & 30.65 & 28.83 & 37.08 & 22.06 & 31.98 & 24.92 & 32.03 & 28.84 & 23.18 & 34.87 & 29.35 \\
& hi & 22.84 & 27.20 & 25.81 & 29.68 & 15.99 & 30.74 & 25.63 & 34.79 & 20.96 & 31.60 & 25.88 & 22.71 & 26.36 & 20.08 & 22.86 & 25.54 \\
& latin & 34.02 & - & 29.34 & 41.53 & 26.43 & 34.90 & 33.45 & 39.81 & 27.31 & 38.11 & 34.81 & 39.27 & 32.54 & 27.49 & 41.81 & 34.34 \\
& dvn. & 32.69 & - & 29.42 & 41.77 & 26.65 & 36.75 & 34.28 & 39.65 & 29.56 & 41.39 & 35.50 & 39.66 & 32.09 & 28.25 & 42.53 & 35.01 \\
& all & 36.14 & 36.15 & 33.70 & 42.06 & 27.40 & 37.25 & 37.80 & 43.62 & 30.09 & 41.68 & 33.08 & 40.69 & 37.44 & 32.49 & 43.69 & 36.88 \\
\midrule
\multirow{5}{*}{\rotatebox[origin=c]{90}{\dataset/-T5}} & en & 32.48 & 20.09 & 32.17 & 43.79 & 26.75 & 33.90 & 33.35 & 39.73 & 25.61 & 27.60 & 33.45 & 35.38 & 31.75 & 26.27 & 36.59 & 31.93 \\
& hi & 29.26 & 29.26 & 33.35 & 38.13 & 28.53 & 40.11 & 35.06 & 41.74 & 29.62 & 34.40 & 35.99 & 40.09 & 33.19 & 28.76 & 41.67 & 34.61 \\
& latin & 32.81 & - & 29.56 & 43.62 & 26.04 & 35.17 & 33.32 & 39.68 & 27.25 & 38.41 & 34.62 & 39.82 & 32.34 & 27.54 & 41.86 & 34.43 \\
& dvn. & 33.99 & - & 30.33 & 43.82 & 28.14 & 40.06 & 35.01 & 38.93 & 33.72 & 44.50 & 36.38 & 40.82 & 31.85 & 28.92 & 44.40 & 36.49 \\
& all & {\bf 41.21} & {\bf 39.17} & {\bf 37.10} & {\bf 43.87} & {\bf 32.00} & {\bf 40.29} & {\bf 41.13} & {\bf 45.16} & {\bf 34.60} & {\bf 44.63} & {\bf 40.56} & {\bf 44.13} & {\bf 39.91} & {\bf 36.69} & {\bf 46.66} & {\bf 40.48} \\
\bottomrule
\end{tabular}
\caption{Headline generation results for the three baseline models trained in all five data settings: English only (en), Hindi only (hi), Latin transliterated data (latin), Devanagari transliterated data (dvn.), and original script data (all). Only \rougel/ scores are shown here. See Appendix \ref{sec:ft-details} for more details.}
\label{tbl:main-results}
\end{table*}

\section{Results}\label{sec:results}
\subsection{Headline Generation Results}
The \rougel/ scores for each of the three models in all five settings can be found in Table \ref{tbl:main-results}. Overall, \dataset/-T5 finetuned in the \textit{all} setting performs the best with an average \rougel/ of 40.48. Although better, it is just six points above the Extractive Oracle baseline. This indicates the difficulty of the \dataset/ dataset, and that there is much room for improvement even among state-of-the-art models.

\subsubsection{mBERT Observations}
On average, mBERT has inferior performance compared to other models, which is expected due to it having to learn a fraction of its parameters from scratch. 
However, when trained on the Devanagari and original scripts, mBERT either outperforms or is comparable to the other models in Bhojpuri and Hindi (both use Devanagari script). It also achieves comparable results to mT5 (with a margin of 0.5 \rougel/ or less) in five languages when trained on the original script data.
But when trained only on English and Hindi data, the model is unable to transfer anything meaningful to other languages, showing that warm starting is not effective in zero-shot cross-lingual settings. This effect is particularly visible in the case of Oriya, which does not share its script with any other language in the dataset, and on which the model has not been pretrained.

However, we observe that when trained only on Hindi data, the model demonstrates faster learning and better transfer capabilities than when trained only on English. This might be because (i) 4/15 languages in the dataset use the Devanagari script, which is also used by Hindi, and (ii) Hindi is closely related to the other languages in the dataset. This suggests that Hindi might be a better bridge language for \dataset/ than English. This hypothesis is further supported by the fact that the Devanagari unified model (\textit{dvn.}) performs significantly better than the Latin unified model.

\subsubsection{mT5 Observations}
\paragraph{Finetuning on Original Scripts}
The model performs the best in this setting, which is expected since it is pretrained on 12 out of 15 languages in the dataset. More importantly, it performs much better than mBERT in Assamese, Bhojpuri, and Oriya, the only three languages both models have not seen during pretraining. This indicates better intra-script and inter-script zero-shot transfer in mT5 (Assamese shares its script with Bengali, and Bhojpuri with Hindi, but Oriya is unique).

\paragraph{Bridge Language}
We find that the mT5 model fine-tuned only on English data demonstrates an average improvement of 3.81 \rougel/ points over the model fine-tuned only on Hindi. This difference in performance can potentially be attributed to the fact that English is the model's largest pretraining language, and it also has the highest share in the vocabulary of the model. Further analysis of the \dataset/-T5 model will support this hypothesis.

\paragraph{Unified Script}
On average, the Devanagari unified model demonstrates better performance, with an improvement of 0.67 \rougel/, when compared to the Latin unified model. Although the performance of the two models is similar for the majority of languages, the Devanagari model shows significant improvement on Marathi, Nepali, and Hindi, all of which also use the Devanagari script. This observation of improved transfer across languages that share a script is a recurrent theme across models and settings. However, it is noteworthy that the Latin model performed better on Assamese and Tamil.

\begin{table}[]
  \centering
  \begin{tabular}{rcccc}
  \toprule
      \multirow{ 2}{*}{Lang.} & \multicolumn{2}{c}{Head. Gen.} & \multicolumn{2}{c}{Abs. Sum.} \\
      \cmidrule(lr){2-3} \cmidrule(lr){4-5}
      & mT5 & V-T5 & mT5 & V-T5 \\
      \midrule 
    bn & 22.32 & \bf{27.02} & 16.12 & \bf{17.68} \\
    en & 19.32 & \bf{21.41} & 13.24 & \bf{14.80} \\
    gu & 14.22 & \bf{17.72} & 10.73 & \bf{12.68} \\
    hi & 19.13 & \bf{23.34} & 18.87 & \bf{22.04} \\
    mr & 16.63 & \bf{20.65} & 9.3 & \bf{10.85} \\
    ne & 15.81 & \bf{20.14} & 9.94 & \bf{12.84} \\
    pa & 19.15 & \bf{23.45} & 15.88 & \bf{18.19} \\
    ta & 19.03 & \bf{22.88} & 12.55 & \bf{14.64} \\
    te & 19.32 & \bf{22.04} & 11.14 & \bf{13.02} \\
    ur & 20.15 & \bf{24.53} & 19.05 & \bf{21.24} \\
      \midrule
    avg. & 18.51 & \bf{22.32} & 13.68 & \bf{15.80} \\
      \bottomrule
  \end{tabular}
  \caption{Zero-shot results on XL-Sum headline generation and abstractive summarization tasks. Only \rougel/ scores are shown here. See Table \ref{tbl:app-xlsum} in the Appendix for more details. }
  \label{tbl:xlsum}
  \end{table}

\subsubsection{\dataset/-T5 observations}
\paragraph{Finetuning on Original Scripts}
\dataset/-T5 fine-tuned in this setting has the highest performance among all models, with a \rougel/ score of 40.48. This result is expected, as it is pretrained on the same data and one of its pretraining objectives (gap-sentence generation) is almost equivalent to generating sentences that the Extractive Oracle would extract. 

\paragraph{Bridge Language}
We see that the model fine-tuned only on Hindi outperforms the model fine-tuned only on English by a margin of 2.68 \rougel/ points. This supports our previous hypothesis that the language with the largest pretraining data is the most effective bridge language. In addition to being the largest language with the largest vocabulary share, Hindi also shares its script with three other languages in the dataset and is typologically similar to the majority of other languages in the dataset. This should ideally make the Hindi-only models significantly better than the English-only models. However, we should note that English is the second biggest language in \dataset/ with 7.2M articles (half the size of Hindi), and therefore the difference in performance between English-only and Hindi-only models, or Latin unified and Devanagari unified models is not as significant as expected.

\paragraph{Unified Script}
On average, the Devanagari unified model is 2.06 \rougel/ points better than the Latin unified model. The Devanagari model performs much better on Gujarati, Hindi, Marathi, Nepali, and Urdu, with improvements of more than 2 \rougel/ points.\footnote{Hindi, Marathi, and Nepali use the Devanagari script.} In general, all Indo-European languages see improvements with the Devanagari unified model. However, such a claim cannot be made for Dravidian languages. While Kannada and Telugu see improvements, we see a performance drop in Malayalam and Telugu.\footnote{Surprisingly, we do not see this drop in mT5 models where English is the dominant pretraining language.} 

\subsection{Generalization Results}
We see that the models pretrained on \dataset/ consistently outperform strong baselines and often by significant margins. We argue (particularly in \S\ref{sec:nlu}) that \dataset/ is a good resource for pretraining large language models on Indic languages, especially since it has data from diverse, high-quality sources.

\subsubsection{XL-Sum}
To test the generalizability of our models on other datasets and tasks, we evaluate models finetuned on \dataset/ on the Indic subset of the XL-Sum dataset \cite{xlsum}, for both abstractive summarization and headline generation. We select the best mT5 and \dataset/-T5 models obtained previously and evaluate them on nine Indic languages and English, without any additional fine-tuning. The results are shown in Table \ref{tbl:xlsum}. We find that \dataset/-T5 consistently outperforms mT5 on all languages in both tasks. On average, we see a gain of around 2 \rougel/ points on abstractive summarization and 4 \rougel/ points on headline generation respectively.

\subsubsection{\dataset/ on NLU tasks}\label{sec:nlu}
To verify if \dataset/ can be used to train good NLU models, we use it to pretrain a masked language model with the BERT-Base architecture. We name this model \dataset/-BERT and more information about the pretraining can be found in Appendix \ref{sec:pt-bert}. We evaluate our model on the IndicXTREME benchmark \cite{indicxtreme} which consists of 9 tasks in 19 Indic languages and English.

We compare our models against two strong baselines: IndicBERT v1 and v2. These two BERT-Base models are trained on IndicCorp v1 \cite{indiccorp} and IndicCorp v2 \cite{indicxtreme}, with 8.5B and 20.9B tokens respectively. We are mainly interested in the comparison with IndicBERT v1 since the size of its pretraining corpus is comparable with that of \dataset/ (9B tokens). Table \ref{tbl:indicxtreme} shows the results averaged across languages on the nine tasks. We see that \dataset/-BERT consistently outperforms IndicBERT v1 indicating its quality. We should also note that \dataset/-BERT's performance is not too far behind IndicBERT v2 even though it is trained on 11B fewer tokens. In fact, it outperforms IndicBERT v2 on three tasks and loses out on three other tasks by a margin of less than one point.

 \begin{table}[]
\centering
\begin{tabular}{r>{\columncolor[gray]{0.9}}ccc}
\toprule
\multirow{2}{*}{Task} & \multicolumn{2}{c}{IndicCorp} & \multirow{2}{*}{\dataset/} \\
\cmidrule(lr){2-3}
& v2 & v1 & \\
 \midrule
Sentiment & 90.2 & 85.7 & \textbf{87.6} \\
NLI   & 73.0 & 66.4 & \textbf{73.6} \\
COPA  & 62.7 & 52.4 & \textbf{60.5} \\
XPara & 56.9 & 49.6 & \textbf{61.9} \\
Intent Clf.   & 78.8 & 25.8 & \textbf{78.0} \\
NER & 73.2 & 58.3 & \textbf{65.2} \\
Slot Fill. & 56.7 & 34.4 & \textbf{57.7} \\
QA    & 48.3 & 37.6 & \textbf{48.3} \\
Retrieval     & 69.4 & \textbf{54.9} & 47.5 \\
\bottomrule
\end{tabular}
\caption{Results on the nine IndicXTREME tasks. Task descriptions, metrics, and detailed results can be found in Appendix \ref{sec:xtreme-app-results}.}
\label{tbl:indicxtreme}
\end{table}

\subsubsection{\dataset/ on NLG tasks}
Finally, we evaluate \dataset/-T5 on IndicNLG and compare its performance against two strong baselines: IndicBART \cite{indicbart} and mT5.\footnote{Here we use the original mT5 model trained on the mC4 corpus.} The comparison is presented in Table \ref{tbl:indicnlg}. It is to be noted that, at the time of writing we could not independently reproduce all the results presented in \citet{indicnlg}. We, therefore, present the results as originally reported along with the ones obtained during our experiments. We see that, overall, \dataset/-T5 is the best-performing model in 3 out of 5 tasks. But when compared with the reproduced results only, it performs better in 4 out of 5 tasks.

\subsection{Key Takeaways}
Based on our experiments and analyses, we now try to answer the research questions posed in \S\ref{sec:rq}.

\paragraph{RQ1} We find that the largest pretraining language typically acts as the best bridge language. It helps if the language is typologically similar to other languages and shares a common script with them. In the case of dataset \dataset/, Hindi is the ideal bridge language as it has all of the above properties (though it shares a common script with only three other languages).

\paragraph{RQ2 and RQ3} The performance of the headline generation models is not negatively impacted by the variety of scripts. In fact, models fine-tuned in this setting yield the best results. It is also interesting to note that the transliterated models perform better than the monolingual models. This suggests that, in general, multilingual training enables positive transfer among related languages, with or without the use of their original scripts.

\paragraph{RQ4} We find that \dataset/-T5 generalizes well on other, similar tasks like abstractive summarization. \dataset/-BERT and \dataset/-T5 generally perform on par or better than strong baselines on both NLU and NLG benchmarks for Indic languages.

\begin{table}[]
\centering
\setlength{\tabcolsep}{5pt} %
\begin{tabular}{lcccc}
\toprule
\multirow{2}{*}{Task} & Indic & \multirow{2}{*}{mT5$^*$} & \multirow{2}{*}{mT5} & \dataset/- \\
& BART$^*$ & & & mT5 \\
\midrule
Wikibio   & 53.8 & \textbf{54.6} & 53.3 & \textbf{54.5} \\
HeadGen.  & 42.4 & \textbf{45.5} & 35.0 & 37.8 \\
SentSum.  & 54.9 & \textbf{55.1} & 36.1 & 32.6 \\
ParaGen.  & 16.2 & 7.5 & 20.5 & \textbf{26.3} \\
QuestGen. & 26.6 & 25.1 & 25.4 & \textbf{28.5} \\
\bottomrule
\end{tabular}
\caption{IndicNLG evaluation results. Columns marked with $^*$ are directly taken from \citet{indicnlg}. More details on tasks and results can be found in Appendix \ref{sec:indicnlg-app-results}.}
\label{tbl:indicnlg}
\end{table}

\section{Conclusion}
In this work, we create \dataset/, a large-scale headline-generation dataset for Indic languages. We show that this dataset is challenging even for state-of-the-art text generation models. Utilizing the size and quality of the dataset, we answer pertinent research questions about multilingual models, from an Indic text generation perspective. We also show that \dataset/ can be used as a pretraining corpus to train strong NLU and NLG models.

\section*{Acknowledgements}
We would like to thank the anonymous reviewers for their comments and suggestions. We also thank Google's TPU Research Cloud (TRC) for giving us free access to their v3-128 TPUs for pretraining our models. We also acknowledge the support of the Natural Sciences and Engineering Research Council of Canada (NSERC) for funding Ziling Cheng through their grant program.

\section*{Limitations}
This work is mainly dedicated to the curation of a new multilingual dataset for Indic languages, many of which are low-resource languages. During data collection, we face several limitations that can potentially result in ethical concerns. Some of the important ones are mentioned below:
\begin{itemize}
\item Our dataset contains only those articles written by DailyHunt's partner publishers. This has the potential to result in a bias towards a particular narrative or ideology that can affect the representativeness and diversity of the dataset.
\item Another limitation is the languages represented in \dataset/. Out of 22 languages with official status in India, our dataset has only 13. There are 122 major languages spoken by at least 10,000 people and 159 other languages which are extremely low-resourced.\footnote{\url{https://en.wikipedia.org/wiki/Languages_of_India}} None of these languages are represented in our dataset.
\item We do not perform any kind of debiasing on \dataset/. This means that societal and cultural biases may exist in the dataset, which can adversely affect the fairness and inclusivity of the models trained on it.
\end{itemize}

\section*{Ethics Statement}
The ethical considerations that arise from the limitations of our data collection process are already detailed in the previous section. In this section, we discuss the implications of releasing the data, how we intend to do so in a safe manner, and the license under which it would be released.

While \dataset/ has the potential to advance NLP research for Indic languages,\footnote{Appendix \ref{sec:datasheet} has a detailed datasheet describing the rationale behind the creation of \dataset/ and other essential information.} it can also be used in ways not intended by the authors. Since \dataset/ can be used to pretrain text generation models, it can be used to build models that generate hate speech, fake news, etc.

Since our data is aggregated from different sources and each source may have different restrictions on distributing their data, we only release a list of URLs pointing to the original articles and not the articles themselves, which is a standard and acceptable way of sharing data.\footnote{Other works like \citet{xsum} also follow this method for sharing their data.} However, we provide a sample script that can be used to crawl the URLs and rebuild \dataset/. We release the URL list under a CC-BY license\footnote{\url{https://creativecommons.org/licenses/by/4.0/}} and dedicate it to the public domain. The released code and models will have an Apache License 2.0.\footnote{\url{https://www.apache.org/licenses/LICENSE-2.0}}

\bibliography{anthology,custom}

\begin{thebibliography}{33}
\expandafter\ifx\csname natexlab\endcsname\relax\def\natexlab#1{#1}\fi

\bibitem[{Aralikatte et~al.(2021)Aralikatte, Narayan, Maynez, Rothe, and
  McDonald}]{aralikatte-etal-2021-focus}
Rahul Aralikatte, Shashi Narayan, Joshua Maynez, Sascha Rothe, and Ryan
  McDonald. 2021.
\newblock \href {https://doi.org/10.18653/v1/2021.acl-long.474} {Focus
  attention: Promoting faithfulness and diversity in summarization}.
\newblock In \emph{Proceedings of the 59th Annual Meeting of the Association
  for Computational Linguistics and the 11th International Joint Conference on
  Natural Language Processing (Volume 1: Long Papers)}, pages 6078--6095,
  Online. Association for Computational Linguistics.

\bibitem[{Banko et~al.(2000)Banko, Mittal, and Witbrock}]{Banko2000HeadlineGB}
Michele Banko, Vibhu Mittal, and M.~Witbrock. 2000.
\newblock Headline generation based on statistical translation.
\newblock In \emph{Annual Meeting of the Association for Computational
  Linguistics}.

\bibitem[{Bhat et~al.(2015)Bhat, Mujadia, Tammewar, Bhat, and
  Shrivastava}]{Bhat:2014:ISS:2824864.2824872}
Irshad~Ahmad Bhat, Vandan Mujadia, Aniruddha Tammewar, Riyaz~Ahmad Bhat, and
  Manish Shrivastava. 2015.
\newblock \href {https://doi.org/10.1145/2824864.2824872} {Iiit-h system
  submission for fire2014 shared task on transliterated search}.
\newblock In \emph{Proceedings of the Forum for Information Retrieval
  Evaluation}, FIRE '14, pages 48--53, New York, NY, USA. ACM.

\bibitem[{Dabre et~al.(2022)Dabre, Shrotriya, Kunchukuttan, Puduppully, Khapra,
  and Kumar}]{indicbart}
Raj Dabre, Himani Shrotriya, Anoop Kunchukuttan, Ratish Puduppully, Mitesh
  Khapra, and Pratyush Kumar. 2022.
\newblock \href {https://doi.org/10.18653/v1/2022.findings-acl.145}
  {{I}ndic{BART}: A pre-trained model for indic natural language generation}.
\newblock In \emph{Findings of the Association for Computational Linguistics:
  ACL 2022}, pages 1849--1863, Dublin, Ireland. Association for Computational
  Linguistics.

\bibitem[{Devlin et~al.(2019)Devlin, Chang, Lee, and
  Toutanova}]{devlin-etal-2019-bert}
Jacob Devlin, Ming-Wei Chang, Kenton Lee, and Kristina Toutanova. 2019.
\newblock \href {https://doi.org/10.18653/v1/N19-1423} {{BERT}: Pre-training of
  deep bidirectional transformers for language understanding}.
\newblock In \emph{Proceedings of the 2019 Conference of the North {A}merican
  Chapter of the Association for Computational Linguistics: Human Language
  Technologies, Volume 1 (Long and Short Papers)}, pages 4171--4186,
  Minneapolis, Minnesota. Association for Computational Linguistics.

\bibitem[{Doddapaneni et~al.(2022)Doddapaneni, Aralikatte, Ramesh, Goyal,
  Khapra, Kunchukuttan, and Kumar}]{indicxtreme}
Sumanth Doddapaneni, Rahul Aralikatte, Gowtham Ramesh, Shreya Goyal, Mitesh~M.
  Khapra, Anoop Kunchukuttan, and Pratyush Kumar. 2022.
\newblock \href {https://doi.org/10.48550/ARXIV.2212.05409} {Indicxtreme: A
  multi-task benchmark for evaluating indic languages}.

\bibitem[{Evans et~al.(2004)Evans, Klavans, and McKeown}]{evans2004columbia}
David~K Evans, Judith~L Klavans, and Kathleen McKeown. 2004.
\newblock Columbia newsblaster: Multilingual news summarization on the web.
\newblock In \emph{Demonstration Papers at HLT-NAACL 2004}, pages 1--4.

\bibitem[{Filippova and Altun(2013)}]{filippova-altun-2013-overcoming}
Katja Filippova and Yasemin Altun. 2013.
\newblock \href {https://aclanthology.org/D13-1155} {Overcoming the lack of
  parallel data in sentence compression}.
\newblock In \emph{Proceedings of the 2013 Conference on Empirical Methods in
  Natural Language Processing}, pages 1481--1491, Seattle, Washington, USA.
  Association for Computational Linguistics.

\bibitem[{Gebru et~al.(2018)Gebru, Morgenstern, Vecchione, Vaughan, Wallach,
  III, and Crawford}]{datasheets}
Timnit Gebru, Jamie Morgenstern, Briana Vecchione, Jennifer~Wortman Vaughan,
  Hanna~M. Wallach, Hal~Daum{\'{e}} III, and Kate Crawford. 2018.
\newblock \href {http://arxiv.org/abs/1803.09010} {Datasheets for datasets}.
\newblock \emph{CoRR}, abs/1803.09010.

\bibitem[{Hasan et~al.(2021)Hasan, Bhattacharjee, Islam, Mubasshir, Li, Kang,
  Rahman, and Shahriyar}]{xlsum}
Tahmid Hasan, Abhik Bhattacharjee, Md.~Saiful Islam, Kazi Mubasshir, Yuan-Fang
  Li, Yong-Bin Kang, M.~Sohel Rahman, and Rifat Shahriyar. 2021.
\newblock \href {https://aclanthology.org/2021.findings-acl.413} {{XL}-sum:
  Large-scale multilingual abstractive summarization for 44 languages}.
\newblock In \emph{Findings of the Association for Computational Linguistics:
  ACL-IJCNLP 2021}, pages 4693--4703, Online. Association for Computational
  Linguistics.

\bibitem[{Kakwani et~al.(2020)Kakwani, Kunchukuttan, Golla, N.C.,
  Bhattacharyya, Khapra, and Kumar}]{indiccorp}
Divyanshu Kakwani, Anoop Kunchukuttan, Satish Golla, Gokul N.C., Avik
  Bhattacharyya, Mitesh~M. Khapra, and Pratyush Kumar. 2020.
\newblock \href {https://doi.org/10.18653/v1/2020.findings-emnlp.445}
  {{I}ndic{NLPS}uite: Monolingual corpora, evaluation benchmarks and
  pre-trained multilingual language models for {I}ndian languages}.
\newblock In \emph{Findings of the Association for Computational Linguistics:
  EMNLP 2020}, pages 4948--4961, Online. Association for Computational
  Linguistics.

\bibitem[{Khanuja et~al.(2022)Khanuja, Ruder, and
  Talukdar}]{evaluating-inclusivity}
Simran Khanuja, Sebastian Ruder, and Partha Talukdar. 2022.
\newblock \href {https://doi.org/10.48550/ARXIV.2205.12676} {Evaluating
  inclusivity, equity, and accessibility of nlp technology: A case study for
  indian languages}.

\bibitem[{Kumar et~al.(2022{\natexlab{a}})Kumar, Shrotriya, Sahu, Dabre,
  Puduppully, Kunchukuttan, Mishra, Khapra, and Kumar}]{indichl}
Aman Kumar, Himani Shrotriya, Prachi Sahu, Raj Dabre, Ratish Puduppully, Anoop
  Kunchukuttan, Amogh Mishra, Mitesh~M. Khapra, and Pratyush Kumar.
  2022{\natexlab{a}}.
\newblock \href {https://doi.org/10.48550/ARXIV.2203.05437} {Indicnlg
  benchmark: Multilingual datasets for diverse nlg tasks in indic languages}.

\bibitem[{Kumar et~al.(2022{\natexlab{b}})Kumar, Shrotriya, Sahu, Dabre,
  Puduppully, Kunchukuttan, Mishra, Khapra, and Kumar}]{indicnlg}
Aman Kumar, Himani Shrotriya, Prachi Sahu, Raj Dabre, Ratish Puduppully, Anoop
  Kunchukuttan, Amogh Mishra, Mitesh~M. Khapra, and Pratyush Kumar.
  2022{\natexlab{b}}.
\newblock \href {https://doi.org/10.48550/ARXIV.2203.05437} {Indicnlg
  benchmark: Multilingual datasets for diverse nlg tasks in indic languages}.

\bibitem[{Kunchukuttan(2020)}]{kunchukuttan2020indicnlp}
Anoop Kunchukuttan. 2020.
\newblock {The IndicNLP Library}.
\newblock
  \url{https://github.com/anoopkunchukuttan/indic_nlp_library/blob/master/docs/indicnlp.pdf}.

\bibitem[{Lewis et~al.(2020)Lewis, Liu, Goyal, Ghazvininejad, Mohamed, Levy,
  Stoyanov, and Zettlemoyer}]{bart}
Mike Lewis, Yinhan Liu, Naman Goyal, Marjan Ghazvininejad, Abdelrahman Mohamed,
  Omer Levy, Veselin Stoyanov, and Luke Zettlemoyer. 2020.
\newblock \href {https://doi.org/10.18653/v1/2020.acl-main.703} {{BART}:
  Denoising sequence-to-sequence pre-training for natural language generation,
  translation, and comprehension}.
\newblock In \emph{Proceedings of the 58th Annual Meeting of the Association
  for Computational Linguistics}, pages 7871--7880, Online. Association for
  Computational Linguistics.

\bibitem[{Lin(2004)}]{lin-2004-rouge}
Chin-Yew Lin. 2004.
\newblock \href {https://aclanthology.org/W04-1013} {{ROUGE}: A package for
  automatic evaluation of summaries}.
\newblock In \emph{Text Summarization Branches Out}, pages 74--81, Barcelona,
  Spain. Association for Computational Linguistics.

\bibitem[{Loshchilov and Hutter(2019)}]{loshchilov2018decoupled}
Ilya Loshchilov and Frank Hutter. 2019.
\newblock \href {https://openreview.net/forum?id=Bkg6RiCqY7} {Decoupled weight
  decay regularization}.
\newblock In \emph{International Conference on Learning Representations}.

\bibitem[{Nallapati et~al.(2017)Nallapati, Zhai, and
  Zhou}]{10.5555/3298483.3298681}
Ramesh Nallapati, Feifei Zhai, and Bowen Zhou. 2017.
\newblock Summarunner: A recurrent neural network based sequence model for
  extractive summarization of documents.
\newblock In \emph{Proceedings of the Thirty-First AAAI Conference on
  Artificial Intelligence}, AAAI'17, page 3075–3081. AAAI Press.

\bibitem[{Nallapati et~al.(2016)Nallapati, Zhou, dos Santos, Gul{\c{c}}ehre,
  and Xiang}]{cnndm}
Ramesh Nallapati, Bowen Zhou, Cicero dos Santos, {\c{C}}a{\u{g}}lar
  Gul{\c{c}}ehre, and Bing Xiang. 2016.
\newblock \href {https://doi.org/10.18653/v1/K16-1028} {Abstractive text
  summarization using sequence-to-sequence {RNN}s and beyond}.
\newblock In \emph{Proceedings of the 20th {SIGNLL} Conference on Computational
  Natural Language Learning}, pages 280--290, Berlin, Germany. Association for
  Computational Linguistics.

\bibitem[{Napoles et~al.(2012)Napoles, Gormley, and
  Van~Durme}]{napoles2012annotated}
Courtney Napoles, Matthew~R Gormley, and Benjamin Van~Durme. 2012.
\newblock Annotated gigaword.
\newblock In \emph{Proceedings of the Joint Workshop on Automatic Knowledge
  Base Construction and Web-scale Knowledge Extraction (AKBC-WEKEX)}, pages
  95--100.

\bibitem[{Narayan et~al.(2018)Narayan, Cohen, and Lapata}]{xsum}
Shashi Narayan, Shay~B. Cohen, and Mirella Lapata. 2018.
\newblock \href {https://doi.org/10.18653/v1/D18-1206} {Don{'}t give me the
  details, just the summary! topic-aware convolutional neural networks for
  extreme summarization}.
\newblock In \emph{Proceedings of the 2018 Conference on Empirical Methods in
  Natural Language Processing}, pages 1797--1807, Brussels, Belgium.
  Association for Computational Linguistics.

\bibitem[{Nenkova(2005)}]{nenkova-2005}
Ani Nenkova. 2005.
\newblock Automatic text summarization of newswire: Lessons learned from the
  document understanding conference.
\newblock pages 1436--1441.

\bibitem[{Ponti et~al.(2021)Ponti, Aralikatte, Shrivastava, Reddy, and
  S{\o}gaard}]{ponti-etal-2021-minimax}
Edoardo~Maria Ponti, Rahul Aralikatte, Disha Shrivastava, Siva Reddy, and
  Anders S{\o}gaard. 2021.
\newblock \href {https://doi.org/10.18653/v1/2021.findings-acl.106} {Minimax
  and neyman{--}{P}earson meta-learning for outlier languages}.
\newblock In \emph{Findings of the Association for Computational Linguistics:
  ACL-IJCNLP 2021}, pages 1245--1260, Online. Association for Computational
  Linguistics.

\bibitem[{Raffel et~al.(2022)Raffel, Shazeer, Roberts, Lee, Narang, Matena,
  Zhou, Li, and Liu}]{t5}
Colin Raffel, Noam Shazeer, Adam Roberts, Katherine Lee, Sharan Narang, Michael
  Matena, Yanqi Zhou, Wei Li, and Peter~J. Liu. 2022.
\newblock Exploring the limits of transfer learning with a unified text-to-text
  transformer.
\newblock \emph{J. Mach. Learn. Res.}, 21(1).

\bibitem[{Rothe et~al.(2020)Rothe, Narayan, and
  Severyn}]{rothe-etal-2020-leveraging}
Sascha Rothe, Shashi Narayan, and Aliaksei Severyn. 2020.
\newblock \href {https://doi.org/10.1162/tacl_a_00313} {Leveraging pre-trained
  checkpoints for sequence generation tasks}.
\newblock \emph{Transactions of the Association for Computational Linguistics},
  8:264--280.

\bibitem[{Rush et~al.(2015)Rush, Chopra, and Weston}]{gigaword}
Alexander~M. Rush, Sumit Chopra, and Jason Weston. 2015.
\newblock \href {https://doi.org/10.18653/v1/D15-1044} {A neural attention
  model for abstractive sentence summarization}.
\newblock In \emph{Proceedings of the 2015 Conference on Empirical Methods in
  Natural Language Processing}, pages 379--389, Lisbon, Portugal. Association
  for Computational Linguistics.

\bibitem[{Scialom et~al.(2020)Scialom, Dray, Lamprier, Piwowarski, and
  Staiano}]{mlsum}
Thomas Scialom, Paul-Alexis Dray, Sylvain Lamprier, Benjamin Piwowarski, and
  Jacopo Staiano. 2020.
\newblock Mlsum: The multilingual summarization corpus.
\newblock \emph{arXiv preprint arXiv:2004.14900}.

\bibitem[{Shazeer and Stern(2018)}]{DBLP:conf/icml/ShazeerS18}
Noam Shazeer and Mitchell Stern. 2018.
\newblock Adafactor: Adaptive learning rates with sublinear memory cost.
\newblock In \emph{{ICML}}, volume~80 of \emph{Proceedings of Machine Learning
  Research}, pages 4603--4611. {PMLR}.

\bibitem[{Shen et~al.(2017)Shen, Lin, Tu, Zhao, Liu, Sun
  et~al.}]{shen2017recent}
Shi-Qi Shen, Yan-Kai Lin, Cun-Chao Tu, Yu~Zhao, Zhi-Yuan Liu, Mao-Song Sun,
  et~al. 2017.
\newblock Recent advances on neural headline generation.
\newblock \emph{Journal of computer science and technology}, 32(4):768--784.

\bibitem[{V{\"o}lske et~al.(2017)V{\"o}lske, Potthast, Syed, and Stein}]{tldr}
Michael V{\"o}lske, Martin Potthast, Shahbaz Syed, and Benno Stein. 2017.
\newblock \href {https://doi.org/10.18653/v1/W17-4508} {{TL};{DR}: Mining
  {R}eddit to learn automatic summarization}.
\newblock In \emph{Proceedings of the Workshop on New Frontiers in
  Summarization}, pages 59--63, Copenhagen, Denmark. Association for
  Computational Linguistics.

\bibitem[{Xue et~al.(2021)Xue, Constant, Roberts, Kale, Al-Rfou, Siddhant,
  Barua, and Raffel}]{xue-etal-2021-mt5}
Linting Xue, Noah Constant, Adam Roberts, Mihir Kale, Rami Al-Rfou, Aditya
  Siddhant, Aditya Barua, and Colin Raffel. 2021.
\newblock \href {https://doi.org/10.18653/v1/2021.naacl-main.41} {m{T}5: A
  massively multilingual pre-trained text-to-text transformer}.
\newblock In \emph{Proceedings of the 2021 Conference of the North American
  Chapter of the Association for Computational Linguistics: Human Language
  Technologies}, pages 483--498, Online. Association for Computational
  Linguistics.

\bibitem[{Zhang et~al.(2020)Zhang, Zhao, Saleh, and Liu}]{pegasus}
Jingqing Zhang, Yao Zhao, Mohammad Saleh, and Peter~J. Liu. 2020.
\newblock Pegasus: Pre-training with extracted gap-sentences for abstractive
  summarization.
\newblock In \emph{Proceedings of the 37th International Conference on Machine
  Learning}, ICML'20. JMLR.org.

\end{thebibliography}
\bibliographystyle{acl_natbib}

\clearpage

\appendix
\section{Implementation Details}\label{sec:impl-details}
\subsection{Finetuning on \dataset/}\label{sec:ft-details}
For all headline generation experiments, we truncate the input article and output text to 512 and 64 tokens respectively, while keeping the output to 128 tokens in summarization experiments. All settings fine-tune the models for 45K steps using an early stopping strategy with a patience of 10. We use NVIDIA RTX8000 GPUs and always maintain an effective batch size of 256. We use the 8-bit AdamW optimizer with a learning rate of 5e-5. We use a linear warmup of 500 steps. During inference, we use beam search with a beam size of 4, a length penalty of 1.0, and constrain the beam so that no trigrams are repeated. The quality of the generated text is evaluated \rouge/ F1-scores \cite{lin-2004-rouge}. We use the multilingual \rouge/ implementation by \citet{xlsum}.

\subsection{Pretraining \dataset/-BERT}\label{sec:pt-bert}
We pretrain the \dataset/-BERT using the standard BERT-Base architecture with 12 encoder layers. We train with a maximum sequence length of 512 tokens with an embedding dimension of 768. We use 12 attention heads with feed-forward width of 3072. To support all the 15 languages in \dataset/ we use a wordpiece vocabulary of size 128K. In total, the model has 184M parameters. The model is trained with AdamW \cite{loshchilov2018decoupled} optimizer with $\alpha=0.9$ and $\beta=0.98$. We use an initial learning rate of 1e-4 with a warm-up of 10K steps and linearly decay the learning rate till the end of training. We train the model for a total of 1M steps which takes 10 days to finish. We use an effective batch size of 4096 and train the model on TPU v3-128 chips.\footnote{\url{https://cloud.google.com/tpu}} 

\subsection{Pretraining \dataset/-T5}\label{sec:pt-t5}
We pretrain \dataset/-T5 using the T5 1.1 base architecture with 12 encoder and decoder layers. We train with maximum sequence lengths of 512 and 256 for the encoder and decoder respectively. We use 12 attention heads with an embedding dimension of 768 and a feed-forward width of 2048. We use a 128K sentencepiece vocabulary. In total, the model has 395M parameters. The model is trained with Adafactor \cite{DBLP:conf/icml/ShazeerS18} optimizer with a warm-up of 10K steps. We use an initial learning rate of 1e-3 and use square root decay till we reach 2M steps. We use an effective batch size of 256 and train the model on TPU v3-8 chips. The model takes 11 days to train.

\section{Misc. Information}
Information about the scripts used by each language in \dataset/ is presented in Table \ref{tbl:app-lang-split}. The table also contains the sizes of the per-language data splits. Table \ref{tbl:app-baseline} contains the full results (\rougeone/, \rougetwo/, and \rougel/) for the extractive baselines. We also show a few random examples of headline-article pairs from \dataset/ in Table \ref{tbl:app-data-example}.

\begin{table*}
\small
\centering
\begin{tabular}{lllllll} 
\toprule
& \multicolumn{2}{c}{Language} & \multicolumn{4}{c}{Number of articles} \\ 
\cmidrule(lr){2-3} \cmidrule(lr){4-7}
Code & Name & Script & Train & Test & Dev & Total \\
\midrule
as & Assamese & Bengali-Assamese & 69,966 & 8,746 & 8,746 & 87,458 \\
bh & Bhojpuri & Devanagari  & 1,244 & 156 & 156 & 1,556 \\
bn & Bengali & Bengali-Assamese & 2,233,029 & 10,000 & 10,000 & 2,253,029 \\
en & English & Latin & 7,251,226 & 10,000 & 10,000 & 7,271,226 \\
gu & Gujarati & Gujarati  & 1,982,243 & 10,000 & 10,000 & 2,002,243 \\
hi & Hindi & Devanagari & 14,420,160 & 10,000 & 10,000 & 14,440,160 \\
kn & Kannada & Kannada-Telugu & 1,446,812 & 10,000 & 10,000 & 1,466,812 \\
ml & Malayalam & Malayalam& 3,447,133 & 10,000 & 10,000 & 3,467,133 \\
mr & Marathi & Devanagari & 2,650,150 & 10,000 & 10,000 & 2,670,150 \\
ne & Nepali & Devanagari  & 26,017 & 3,253 & 3,253 & 32,523 \\
or & Oriya & Oriya & 1,072,984 & 10,000 & 10,000 & 1,092,984 \\
pa & Punjabi & Gurmukhī & 822,316 & 10,000 & 10,000 & 842,316 \\
ta & Tamil & Tamil & 2,620,616 & 10,000 & 10,000 & 2,640,616 \\
te & Telugu & Kannada-Telugu & 3,254,377 & 10,000 & 10,000 & 3,274,377 \\
ur & Urdu & Urdu & 283,857 & 10,000 & 10,000 & 303,857 \\
\bottomrule
\end{tabular}
\caption{ISO 639-1 language codes, language names, and their corresponding written scripts, sizes of train, validation, and test splits for each language in \dataset/.}
\label{tbl:app-lang-split}
\end{table*}

\begin{table*}
\small
\centering
\begin{tabular}{rccccccccc} 
\toprule
\multirow{2}{*}{Lang.} & \multicolumn{3}{c}{Lead-1} &  \multicolumn{3}{c}{Lead-2} &  \multicolumn{3}{c}{Ext Or.} \\
\cmidrule(lr){2-4} \cmidrule(lr){5-7} \cmidrule(lr){8-10}
& R1 &  R2 & RL & R1 &  R2 & RL & R1 &  R2 & RL \\
\midrule
as & 20.05 & 10.21 & 18.41 & 18.57 & 8.99 & 16.35 & 38.53 & 25.12 & 35.57 \\
bh & 31.76 & 13.58 & 27.11 & 28.02 & 11.49 & 23.64 & 40.86 & 21.27 & 35.26 \\
bn & 22.01 & 11.3 & 20.2 & 20.85 & 10.04 & 18.44 & 39.55 & 23.05 & 35.62 \\
en & 28.37 & 12.86 & 23.78 & 23.45 & 10.44 & 19.25 & 38.23 & 20.76 & 32.83 \\
gu & 16.07 & 6.98 & 14.59 & 17.06 & 7.35 & 15.11 & 33.48 & 18.83 & 30.55 \\
hi & 20.88 & 9.22 & 17.71 & 26.81 & 11.72 & 22.12 & 42.36 & 23.14 & 35.85 \\
kn & 20.86 & 9.47 & 19.36 & 19.11 & 8.35 & 17.34 & 34.91 & 19.01 & 32.4 \\
ml & 33.74 & 19.86 & 32.41 & 26.83 & 14.84 & 24.88 & 43.75 & 28.38 & 42.07 \\
mr & 15.67 & 6.88 & 14.53 & 15.59 & 6.53 & 14.15 & 33.86 & 18.54 & 31.33 \\
ne & 4.61 & 1.44 & 4.25 & 25.09 & 12.46 & 23.35 & 38.77 & 21.55 & 36.38 \\
or & 23.28 & 11.38 & 21.52 & 21.36 & 9.92 & 19.19 & 39.1 & 22.24 & 35.89 \\
pa & 25.53 & 12.83 & 22.24 & 22.54 & 10.96 & 19.18 & 36.63 & 20.26 & 31.75 \\
ta & 25.36 & 12.63 & 23.77 & 19.27 & 9.22 & 17.75 & 36.65 & 20.47 & 34.23 \\
te & 17.69 & 7.52 & 16.67 & 14.56 & 5.9 & 13.42 & 32.25 & 16.72 & 30.31 \\
ur & 29.93 & 16.01 & 25.91 & 23.58 & 12.4 & 20.04 & 39.95 & 24.31 & 35.08 \\
 \midrule
Average & 22.39&	10.81&	20.16&	21.51&10.04&	18.95&		37.93&	21.57&	34.34\\
\bottomrule
\end{tabular}
\caption{Lead-1, Lead-2, and Extractive Oracle performance on \dataset/ test sets.}
\label{tbl:app-baseline}
\end{table*}

\begin{table*}
\small
\centering
\includegraphics[width=\textwidth]{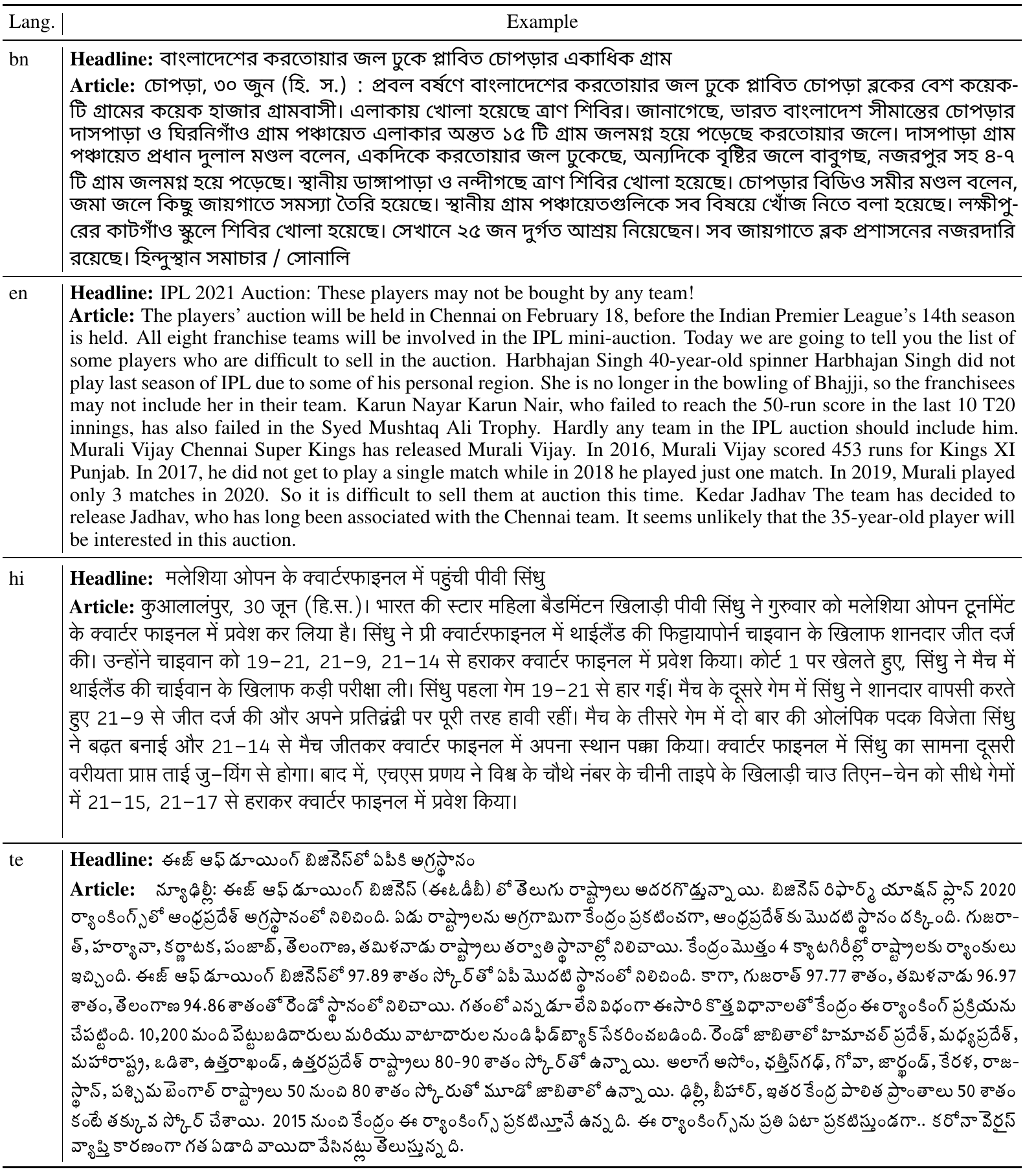} 
\caption{Random examples from \dataset/.}
\label{tbl:app-data-example}
\end{table*}

\section{Finetuning Results}\label{sec:ft-results}
We present detailed results of our headline generation finetuning experiments in Table \ref{tbl:app-mBERT}, \ref{tbl:app-mT5}, and \ref{tbl:app-Varta-T5}. For each model (mBERT, mT5, and \dataset/-T5), we present the \rougeone/, \rougetwo/, and \rougel/ F1-scores in all five settings: \textit{en}, \textit{hi}, \textit{latin}, \textit{dvn.}, and \textit{\textit{all}}.

In Table \ref{tbl:app-xlsum}, we present the zero-shot results of mT5 and \dataset/-T5 on both headline generation and abstractive summarization tasks of the XL-Sum dataset.

\begin{table*}
\small
\centering
\setlength{\tabcolsep}{3pt}
\begin{tabular}{rccccccccccccccc} 
\toprule
\multirow{2}{*}{Lang.} & \multicolumn{3}{c}{en} & \multicolumn{3}{c}{hi}  &  \multicolumn{3}{c}{latin} &  \multicolumn{3}{c}{dev.} & \multicolumn{3}{c}{all} \\
\cmidrule(lr){2-4} \cmidrule(lr){5-7} \cmidrule(lr){8-10} \cmidrule(lr){11-13} \cmidrule(lr){14-16}
 &R1 &  R2 & RL&  R1 &  R2 & RL & R1 &  R2 & RL & R1 &  R2 & RL &  R1 &  R2 & RL\\
\midrule
as & 1.28 & 0.47 & 1.24 & 1.43 & 0.85 & 1.41 & 29.32 & 17.75 & 28.04 & 32.09 & 19.22 & 30.66 & 34.70 & 20.85 & 32.98 \\
bh & 3.09 & 0.66 & 2.82 & 34.47 & 14.84 & 29.86 & - & - & - & - & - & - & 41.96 & 23.91 & 37.20 \\
bn & 2.00 & 0.69 & 1.92 & 1.37 & 0.76 & 1.35 & 26.60 & 13.84 & 24.92 & 29.82 & 16.41 & 28.24 & 36.32 & 19.42 & 33.68 \\
en & 43.96 & 24.15 & 39.14 & 16.73 & 9.04 & 15.50 & 39.41 & 20.93 & 34.89 & 43.80 & 24.94 & 39.37 & 46.21 & 26.81 & 41.36 \\
gu & 1.73 & 0.45 & 1.70 & 1.36 & 0.45 & 1.35 & 22.55 & 10.31 & 21.15 & 26.32 & 13.01 & 24.88 & 28.59 & 14.24 & 26.81 \\
hi & 4.83 & 1.38 & 4.44 & 45.03 & 24.80 & 39.89 & 32.61 & 15.52 & 28.98 & 38.87 & 20.12 & 34.55 & 41.89 & 22.07 & 36.89 \\
kn & 2.83 & 0.70 & 2.73 & 1.36 & 0.47 & 1.35 & 29.66 & 14.67 & 28.10 & 33.47 & 17.20 & 32.09 & 37.10 & 20.02 & 35.50 \\
ml & 1.80 & 0.40 & 1.75 & 1.02 & 0.30 & 1.01 & 36.54 & 22.89 & 35.52 & 38.11 & 23.28 & 37.13 & 41.09 & 26.17 & 39.88 \\
mr & 2.95 & 1.12 & 2.88 & 20.21 & 9.43 & 19.35 & 22.73 & 10.16 & 21.53 & 28.04 & 14.15 & 26.91 & 30.68 & 15.79 & 29.29 \\
ne & 2.36 & 0.25 & 2.29 & 31.27 & 15.41 & 29.74 & 32.42 & 16.25 & 31.14 & 40.40 & 22.70 & 38.80 & 43.41 & 25.39 & 41.55 \\
or & 0.36 & 0.12 & 0.35 & 0.27 & 0.18 & 0.26 & 29.90 & 15.25 & 28.08 & 35.22 & 18.63 & 33.29 & 1.25 & 0.46 & 1.24 \\
pa & 2.55 & 0.56 & 2.44 & 1.93 & 0.54 & 1.89 & 35.50 & 19.47 & 32.52 & 40.64 & 23.62 & 37.48 & 44.42 & 26.10 & 40.60 \\
ta & 2.11 & 0.38 & 2.04 & 1.08 & 0.23 & 1.07 & 28.61 & 14.83 & 27.36 & 31.82 & 16.23 & 30.77 & 37.70 & 21.40 & 36.25 \\
te & 3.49 & 1.27 & 3.42 & 2.38 & 1.21 & 2.36 & 25.02 & 12.30 & 24.20 & 28.34 & 14.06 & 27.40 & 31.18 & 16.26 & 30.21 \\
ur & 2.92 & 1.11 & 2.75 & 2.21 & 1.00 & 2.17 & 38.97 & 22.63 & 35.72 & 43.75 & 27.18 & 40.62 & 47.39 & 29.95 & 43.82 \\

\midrule
Average & 5.22 & 2.25 & 4.79 & 10.81 & 5.30 & 9.90 & 30.70 & 16.20 & 28.72 & 35.05 & 19.34 & 33.01 & 36.26 & 20.59 & 33.82 \\
\bottomrule
\end{tabular}
\caption{Performance of mBERT on \dataset/ headline generation.}
\label{tbl:app-mBERT}
\end{table*}

\begin{table*}
\small
\centering
\setlength{\tabcolsep}{3pt}
\begin{tabular}{rccccccccccccccc} 
\toprule
\multirow{2}{*}{Lang.} & \multicolumn{3}{c}{en} & \multicolumn{3}{c}{hi}&  \multicolumn{3}{c}{latin} &  \multicolumn{3}{c}{dev.}& \multicolumn{3}{c}{all} \\
\cmidrule(lr){2-4} \cmidrule(lr){5-7} \cmidrule(lr){8-10} \cmidrule(lr){11-13} \cmidrule(lr){14-16}
 &R1 &  R2 & RL&  R1 &  R2 & RL & R1 &  R2 & RL & R1 &  R2 & RL &  R1 &  R2 & RL \\
\midrule
as & 25.78 & 14.70 & 24.19 & 24.29 & 13.41 & 22.84 & 35.97 & 23.39 & 34.02 & 34.60 & 21.42 & 32.69 & 38.76 & 22.60 & 36.14 \\
bh & 33.43 & 14.52 & 29.01 & 31.97 & 12.54 & 27.20 & - & - & - & - & - & - & 41.55 & 22.77 & 36.15 \\
bn & 30.92 & 15.86 & 27.73 & 28.49 & 14.48 & 25.81 & 31.56 & 17.66 & 29.34 & 31.50 & 17.71 & 29.42 & 36.65 & 20.08 & 33.70 \\
en & 48.43 & 28.60 & 43.20 & 34.89 & 17.26 & 29.68 & 46.84 & 27.06 & 41.53 & 47.16 & 27.40 & 41.77 & 47.45 & 27.65 & 42.06 \\
gu & 23.24 & 11.31 & 21.66 & 16.91 & 7.13 & 15.99 & 28.57 & 14.58 & 26.43 & 28.65 & 14.59 & 26.65 & 29.44 & 14.96 & 27.40 \\
hi & 35.62 & 17.30 & 30.65 & 35.62 & 17.49 & 30.74 & 39.68 & 20.78 & 34.90 & 41.99 & 22.36 & 36.75 & 42.55 & 22.75 & 37.25 \\
kn & 30.60 & 16.01 & 28.83 & 27.42 & 13.97 & 25.63 & 35.12 & 18.92 & 33.45 & 36.03 & 19.49 & 34.28 & 39.60 & 22.31 & 37.80 \\
ml & 38.77 & 26.39 & 37.08 & 36.68 & 24.65 & 34.79 & 41.65 & 27.88 & 39.81 & 41.32 & 28.36 & 39.65 & 45.45 & 32.22 & 43.62 \\
mr & 23.40 & 11.42 & 22.06 & 22.40 & 10.44 & 20.96 & 28.87 & 14.72 & 27.31 & 31.19 & 16.36 & 29.56 & 31.71 & 16.67 & 30.09 \\
ne & 33.71 & 17.56 & 31.98 & 33.72 & 17.49 & 31.60 & 40.15 & 22.10 & 38.11 & 43.63 & 25.32 & 41.39 & 43.86 & 25.47 & 41.68 \\
or & 26.96 & 12.58 & 24.92 & 27.79 & 13.21 & 25.88 & 37.24 & 20.77 & 34.81 & 37.99 & 21.17 & 35.50 & 35.29 & 18.94 & 33.08 \\
pa & 35.71 & 19.73 & 32.03 & 24.47 & 11.00 & 22.71 & 43.34 & 25.51 & 39.27 & 43.75 & 25.74 & 39.66 & 44.79 & 26.65 & 40.69 \\
ta & 30.49 & 16.35 & 28.84 & 28.06 & 14.56 & 26.36 & 34.14 & 18.82 & 32.54 & 33.58 & 18.27 & 32.09 & 39.09 & 22.54 & 37.44 \\
te & 24.48 & 12.09 & 23.18 & 21.30 & 9.84 & 20.08 & 28.81 & 14.90 & 27.49 & 29.52 & 15.18 & 28.25 & 33.68 & 18.59 & 32.49 \\
ur & 38.10 & 22.40 & 34.87 & 24.47 & 10.59 & 22.86 & 45.79 & 28.52 & 41.81 & 46.49 & 29.20 & 42.53 & 47.57 & 30.15 & 43.69 \\

\midrule
Average & 31.98 & 17.12 & 29.35 & 27.90 & 13.87 & 25.54 & 36.98 & 21.12 & 34.34 & 37.67 & 21.61 & 35.01 & 39.83 & 22.96 & 36.88 \\
\bottomrule
\end{tabular}
\caption{Performance of mT5 on \dataset/ headline generation.}
\label{tbl:app-mT5}
\end{table*}

\begin{table*}
\small
\centering
\setlength{\tabcolsep}{3pt}
\begin{tabular}{rccccccccccccccc} 
\toprule
\multirow{2}{*}{Lang.} & \multicolumn{3}{c}{en} & \multicolumn{3}{c}{hi} & \multicolumn{3}{c}{latin} &  \multicolumn{3}{c}{dev.} & \multicolumn{3}{c}{all} \\ 
\cmidrule(lr){2-4} \cmidrule(lr){5-7} \cmidrule(lr){8-10} \cmidrule(lr){11-13} \cmidrule(lr){14-16}
 & R1 &  R2 & RL&  R1 &  R2 & RL & R1 &  R2 & RL & R1 &  R2 & RL &  R1 &  R2 & RL\\
\midrule
as & 35.01 & 22.61 & 32.48 & 34.45 & 15.82 & 29.26 & 34.69 & 21.99 & 32.81 & 35.83 & 22.51 & 33.99 & 43.22 & 29.47 & 41.21 \\
bh & 23.88 & 9.49 & 20.09 & 34.45 & 15.82 & 29.26 & - & - & - & - & - & - & 44.44 & 24.98 & 39.17 \\
bn & 35.71 & 19.32 & 32.17 & 36.68 & 19.97 & 33.35 & 31.75 & 17.78 & 29.56 & 32.32 & 18.34 & 30.33 & 40.12 & 22.65 & 37.10 \\
en & 49.13 & 29.20 & 43.79 & 43.78 & 24.09 & 38.13 & 48.99 & 28.93 & 43.62 & 49.15 & 29.09 & 43.82 & 49.15 & 29.27 & 43.87 \\
gu & 29.03 & 15.63 & 26.75 & 30.68 & 16.60 & 28.53 & 28.09 & 14.19 & 26.04 & 30.13 & 15.60 & 28.14 & 34.16 & 18.81 & 32.00 \\
hi & 39.64 & 20.50 & 33.90 & 45.45 & 25.40 & 40.11 & 39.96 & 21.03 & 35.17 & 45.33 & 25.24 & 40.06 & 45.64 & 25.56 & 40.29 \\
kn & 35.65 & 19.82 & 33.35 & 37.17 & 20.85 & 35.06 & 35.06 & 18.63 & 33.32 & 36.68 & 19.93 & 35.01 & 42.93 & 25.35 & 41.13 \\
ml & 41.48 & 28.07 & 39.73 & 43.58 & 29.71 & 41.74 & 41.44 & 27.70 & 39.68 & 40.49 & 26.93 & 38.93 & 46.72 & 32.72 & 45.16 \\
mr & 27.38 & 14.70 & 25.61 & 31.47 & 17.13 & 29.62 & 28.81 & 14.45 & 27.25 & 35.42 & 19.86 & 33.72 & 36.29 & 20.58 & 34.60 \\
ne & 29.72 & 14.56 & 27.60 & 36.71 & 19.57 & 34.40 & 40.40 & 22.50 & 38.41 & 46.52 & 27.92 & 44.50 & 46.68 & 27.91 & 44.63 \\
or & 36.46 & 19.90 & 33.45 & 38.68 & 21.72 & 35.99 & 37.04 & 20.58 & 34.62 & 38.79 & 21.97 & 36.38 & 43.09 & 25.61 & 40.56 \\
pa & 39.80 & 22.92 & 35.38 & 44.12 & 26.46 & 40.09 & 43.87 & 25.95 & 39.82 & 44.78 & 26.64 & 40.82 & 48.19 & 29.78 & 44.13 \\
ta & 33.77 & 18.40 & 31.75 & 35.08 & 19.44 & 33.19 & 33.91 & 18.67 & 32.34 & 33.35 & 18.08 & 31.85 & 41.56 & 24.42 & 39.91 \\
te & 27.77 & 14.86 & 26.27 & 30.21 & 16.60 & 28.76 & 28.79 & 14.59 & 27.54 & 30.11 & 15.77 & 28.92 & 37.91 & 22.34 & 36.69 \\
ur & 40.76 & 24.82 & 36.59 & 45.78 & 28.80 & 41.67 & 45.77 & 28.49 & 41.86 & 48.36 & 30.84 & 44.40 & 50.59 & 33.04 & 46.66 \\
\midrule
Average & 35.01 & 19.65 & 31.93 & 37.89 & 21.20 & 34.61 & 37.04 & 21.11 & 34.43 & 39.09 & 22.77 & 36.49 & 43.38 & 26.17 & 40.48 \\
\bottomrule
\end{tabular}
\caption{Performance of \dataset/-T5 on \dataset/ headline generation.}
\label{tbl:app-Varta-T5}
\end{table*}

\begin{table*}
\small
\centering
\begin{tabular}{rcccccccccccc} 
\toprule
& \multicolumn{6}{c}{Headline Generation} & \multicolumn{6}{c}{Abstractive Summarization} \\
\cmidrule(lr){1-7} \cmidrule(lr){8-13}
 \multirow{2}{*}{Lang.} & \multicolumn{3}{c}{mT5} & \multicolumn{3}{c}{\dataset/-T5} & \multicolumn{3}{c}{mT5 } &  \multicolumn{3}{c}{\dataset/-T5} \\
  \cmidrule(lr){2-4} \cmidrule(lr){5-7} \cmidrule(lr){8-10} \cmidrule(lr){11-13}
 &R1 &  R2 & RL&  R1 &  R2 & RL & R1 &  R2 & RL & R1 &  R2 & RL  \\
\midrule
bn & 24.86 & 8.89 & 22.32 & 29.81 & 11.19 & 27.02 & 18.60 & 6.18 & 16.12 & 20.25 & 6.90 & 17.68 \\
en & 22.56 & 6.51 & 19.32 & 25.00 & 7.62 & 21.41 & 16.79 & 3.07 & 13.24 & 18.54 & 3.90 & 14.80 \\
gu & 15.57 & 4.70 & 14.22 & 19.36 & 6.43 & 17.72 & 12.17 & 3.16 & 10.73 & 14.22 & 4.17 & 12.68 \\
hi & 22.43 & 6.39 & 19.13 & 26.81 & 8.89 & 23.34 & 24.66 & 6.85 & 18.87 & 27.60 & 9.24 & 22.04 \\
mr & 18.03 & 7.02 & 16.63 & 22.28 & 9.05 & 20.65 & 10.29 & 3.17 & 9.30 & 11.87 & 3.72 & 10.85 \\
ne & 16.84 & 4.93 & 15.81 & 21.49 & 7.18 & 20.14 & 11.10 & 2.41 & 9.94 & 14.15 & 3.80 & 12.84 \\
pa & 22.29 & 7.42 & 19.15 & 26.92 & 10.12 & 23.45 & 19.65 & 6.05 & 15.88 & 22.31 & 7.29 & 18.19 \\
ta & 20.39 & 8.08 & 19.03 & 24.60 & 10.05 & 22.88 & 13.65 & 4.66 & 12.55 & 16.01 & 5.57 & 14.64 \\
te & 20.90 & 8.01 & 19.32 & 23.85 & 9.51 & 22.04 & 12.24 & 3.44 & 11.14 & 14.27 & 4.48 & 13.02 \\
ur & 22.34 & 7.63 & 20.15 & 26.87 & 10.57 & 24.53 & 24.04 & 7.96 & 19.05 & 25.74 & 9.79 & 21.24 \\
\midrule
Average & 20.62 & 6.96 & 18.51 & 24.70 & 9.06 & 22.32 & 16.32 & 4.70 & 13.68 & 18.50 & 5.89 & 15.80 \\
\bottomrule
\end{tabular}
\caption{Zero-shot performance of the best mT5 and \dataset/-T5 models on XL-Sum headline generation and abstractive summarization.}
\label{tbl:app-xlsum}
\end{table*}

\section{IndicXTREME Results}\label{sec:xtreme-app-results}
IndicXTREME is a cross-lingual benchmark consisting of nine tasks in 18 Indic languages. The nine tasks are as follows: (i) IndicCOPA: for commonsense causal reasoning, (ii) IndicQA: for question-answering, (iii) IndicXParaphrase: for paraphrase detection, (iv) IndicSentiment: for sentiment classification, (v) IndicXNLI: for natural language inference, (vi) Naamapadam: for named entity recognition, (vii) MASSIVE: for intent classification and slot filling (also classification), and (ix) FLORES: for retrieval using semantic similarity.

\dataset/-BERT is compared with two other BERT models pretrained on IndicCorp v1 and v2 respectively. For each task, we finetune the three models on English training data and evaluate them on Indic languages. For a comprehensive list of training data, please refer \citet{indicxtreme}.

For every task, the per-language results for all three models are presented in Tables \ref{tbl:app-indicsentiment}, \ref{tbl:indicXNLI}, \ref{tbl:app-indiccopa}, \ref{tbl:app-indicxparaphrase}, \ref{tbl:app-massive-intent-classification}, \ref{tbl:app-naamapadam}, \ref{tbl:massive slot-filling}, \ref{tbl:app-indicQA}, and \ref{tbl:app-flores} respectively.

\begin{table*}[]
\centering
\setlength{\tabcolsep}{5pt}
\begin{tabular}{rcccccccccccccc}
\toprule
Corpus & as & bd & bn & gu & hi & kn & ml & mr & or & pa & ta & te & ur & Avg. \\
\midrule
IC v1 & 90.9 & 60.2 & 92.7 & 91.9 & 92.2 & 90.6 & 90.1 & 91.9 & 88.2 & 90.6 & 90.6 & 91.6 & 52.9 & 85.7 \\
IC v2 & 91.4 & 80.4 & 91.8 & 90.5 & 91.4 & 90.1 & 90.3 & 91.7 & 90.7 & 91.6 & 92.3 & 91.6 & 89.0 & 90.2 \\
\dataset/ & 90.7 & 50 & 92.6 & 90.7 & 92.7 & 88.6 & 90.5 & 91.7 & 89.9 & 90.9 & 91.9 & 89.3 & 89.2 & 87.6 \\
\bottomrule
\end{tabular}
\caption{Results on IndicSentiment (IndicXTREME). Metric: Accuracy}
\label{tbl:app-indicsentiment}
\end{table*}

\begin{table*}[]
\centering
\setlength{\tabcolsep}{5pt}
\begin{tabular}{rccccccccccccc}
\toprule
Corpus & as & bn & gu & hi & kn & ml & mr & or & pa & ta & te & ur & Avg. \\
\midrule
IC v1 & 67.0 & 70.4 & 70.4 & 72.3 & 69.6 & 67.5 & 68.2 & 69.0 & 71.1 & 68.5 & 68.6 & 34.0 & 66.4 \\
IC v2 & 70.4 & 74.3 & 74.4 & 76.0 & 73.8 & 73.9 & 72.1 & 72.6 & 76.2 & 73.9 & 72.9 & 65.7 & 73.0 \\
\dataset/ & 71.1 & 75.2 & 74.8 & 77.1 & 74.7 & 73.7 & 71.8 & 73.3 & 76.2 & 74.1 & 73.8 & 67.0 & 73.6 \\
\bottomrule
\end{tabular}
\caption{Results on IndicXNLI (IndicXTREME). Metric: Accuracy}
\label{tbl:indicXNLI}
\end{table*}

\begin{table*}[]
\centering
\begin{tabular}{rcccccccccc}
\toprule
Corpus & as & bn & gom & gu & hi & kn & mai & ml & mr & ne \\ 
\midrule
IC v1 & 54.8 & 52.0 & 47.8 & 53.6 & 50.8 & 50.8 & 47.6 & 54.2 & 53.5 & 53.0 \\
IC v2 & 61.2 & 68.8 & 58.2 & 63.2 & 62.4 & 65.8 & 61.2 & 62.6 & 63.7 & 63.0 \\
\dataset/ & 64.2 & 64.0 & 59.4 & 65.6 & 64.6 & 66.4 & 55.6 & 62.6 & 62.1 & 60.8 \\ 
\midrule
 &  & or & pa & sa & sat & sd & ta & te & ur & Avg. \\
 \midrule
IC v1 &  & 53.8 & 55.0 & 47.0 & 50.6 & 53.0 & 54.8 & 50.8 & 55.0 & 52.4 \\
IC v2 &  & 62.8 & 67.0 & 57.6 & 48.2 & 59.2 & 67.2 & 65.4 & 64.8 & 62.7 \\
\dataset/ &  & 64.0 & 61.8 & 49.2 & 46.2 & 51.0 & 61.6 & 67.4 & 60.0 & 60.5 \\
\bottomrule
\end{tabular}
\caption{Results on IndicCOPA (IndicXTREME). Metric: Accuracy.}
\label{tbl:app-indiccopa}
\end{table*}

\begin{table*}[]
\centering
\setlength{\tabcolsep}{5pt}
\begin{tabular}{lccccccccccc}
\toprule
Corpus & as & bn & gu & hi & kn & ml & mr & or & pa & te & Avg. \\
\midrule
IC v1 & 49.5 & 49.5 & 52.6 & 49.2 & 48.0 & 49.1 & 47.9 & 49.6 & 51.2 & 49.5 & 49.6 \\
IC v2 & 57.1 & 50.1 & 74.9 & 50.3 & 57.9 & 56.8 & 54.3 & 57.2 & 55.0 & 55.2 & 56.9 \\
\dataset/ & 62.5 & 52.3 & 82.1 & 54.9 & 62.5 & 61.5 & 64.1 & 62.6 & 55.3 & 61.6 & 61.9 \\
\bottomrule
\end{tabular}
\caption{Results on IndicXParaphrase (IndicXTREME). Metric: Accuracy.}
\label{tbl:app-indicxparaphrase}
\end{table*}

\begin{table*}[]
\centering
\begin{tabular}{lcccccccc}
\toprule
Corpus & bn & hi & kn & ml & ta & te & ur & Avg. \\
\midrule
IC v1 & 31.3 & 32.9 & 30.0 & 29.7 & 25.5 & 30.5 & 1.1 & 25.8 \\
IC v2 & 79.5 & 82.7 & 78.2 & 80.4 & 76.1 & 77.9 & 76.9 & 78.8 \\
\dataset/ & 78.6 & 81.7 & 76.6 & 79.3 & 76.5 & 77.6 & 75.6 & 78.0 \\
\bottomrule
\end{tabular}
\caption{Results on MASSIVE Intent Classification (IndicXTREME). Metric: Accuracy.}
\label{tbl:app-massive-intent-classification}
\end{table*}

\begin{table*}[]
\centering
\begin{tabular}{lcccccccccc}
\toprule
Corpus & bn & gu & hi & kn & ml & mr & pa & ta & te & Avg. \\
\midrule
IC v1 & 60.7 & 58.6 & 61.9 & 58.4 & 60.1 & 53.1 & 55.1 & 51.3 & 65.4 & 58.3 \\
IC v2 & 74.1 & 72.5 & 78.5 & 74.8 & 72.5 & 71.7 & 71.4 & 63.7 & 79.8 & 73.2 \\
\dataset/ & 65.6 & 66.8 & 66.6 & 67.9 & 67.8 & 64.2 & 61.5 & 54.5 & 72.2 & 65.2 \\
\bottomrule
\end{tabular}
\caption{Results on Naamapadam (IndicXTREME). Metric: F1 score.}
\label{tbl:app-naamapadam}
\end{table*}

\begin{table*}[]
\centering
\begin{tabular}{lcccccccc}
\toprule
Corpus & bn & hi & kn & ml & ta & te & ur & Avg. \\
\midrule
IC v1 & 41.1 & 42.8 & 42.2 & 38.6 & 34.4 & 40.6 & 0.8 & 34.4 \\
IC v2 & 61.6 & 55.4 & 55.9 & 60.4 & 56.8 & 58.3 & 48.5 & 56.7 \\
\dataset/ & 60.0 & 58.6 & 58.0 & 61.0 & 55.9 & 58.9 & 51.5 & 57.7 \\
\bottomrule
\end{tabular}
\caption{Results on MASSIVE slot-filling (IndicXTREME). Metric: F1 score.}
\label{tbl:massive slot-filling}
\end{table*}

\begin{table*}[]
\centering
\begin{tabular}{lcccccccccccc}
\toprule
Corpus & as & bn & gu & hi & kn & ml & mr & or & pa & ta & te & Avg. \\
\midrule
IC v1 & 30.8 & 39.7 & 35.8 & 37.7 & 34.7 & 36.2 & 38.9 & 37.6 & 39.8 & 34.4 & 48.1 & 37.6 \\
IC v2 & 44.5 & 51.6 & 43.8 & 54.7 & 45.9 & 43.7 & 46.3 & 47.2 & 51.1 & 43.5 & 59.1 & 48.3 \\
\dataset/ & 44.4 & 52.3 & 44.8 & 55.0 & 44.7 & 41.1 & 47.5 & 48.5 & 51.0 & 44.6 & 57.2 & 48.3 \\
\bottomrule
\end{tabular}
\caption{Results on IndicQA (IndicXTREME). Metric: F1 score.}
\label{tbl:app-indicQA}
\end{table*}

\begin{table*}[]
\centering
\begin{tabular}{lcccccccccc}
\toprule
Corpus & as & bn & gu & hi & kn & ks & mai & ml & mr & mni \\
\midrule
IC v1 & 77.7 & 85.6 & 89.6 & 89.8 & 84.5 & 0.6 & 23.4 & 80.2 & 87.9 & 1.9 \\
IC v2 & 86.0 & 91.0 & 92.4 & 90.5 & 89.1 & 0.9 & 38.1 & 89.2 & 92.5 & 0.3 \\
\dataset/ & 35.4 & 67.3 & 72.9 & 74.7 & 70.4 & 5.6 & 17.4 & 47.0 & 75.0 & 1.4 \\
\midrule
 &  & ne & or & pa & sa & sat & ta & te & ur & Avg. \\
\midrule
IC v1 &  & 16.0 & 82.9 & 88.3 & 9.5 & 0.7 & 83.9 & 84.7 & 0.2 & 54.9 \\
IC v2 &  & 79.9 & 90.9 & 92.2 & 30.4 & 19.9 & 90.0 & 88.6 & 87.0 & 69.4 \\
\dataset/ &  & 50.6 & 60.9 & 67.0 & 7.4 & 0.0 & 74.3 & 62.9 & 64.6 & 47.5 \\
\bottomrule
\end{tabular}
\caption{Results on FLORES (IndicXTREME). Metric: Accuracy.}
\label{tbl:app-flores}
\end{table*}

\section{IndicNLG Results}\label{sec:indicnlg-app-results}
IndicNLG is a benchmark for evaluating NLG in 11 Indic languages. It consists of five tasks: (i) generate the first sentence of a Wikipedia page, given its infobox, (ii) headline generation, (iii) sentence summarization, (iv) paraphrase generation, and (v) question generation. More information about the datasets can be found in \citet{indicnlg}.

We compare \dataset/-T5 with IndicBART and mT5 results reported in \cite{indicnlg}, and also with a version of mT5 finetuned by us. All tasks use \rouge/ as their metric, except paraphrase generation which uses BLEU. The task--wise results on all supported languages for mT5 and \dataset/-T5 are presented in Tables \ref{tbl:app-indicNLG-mt5} and \ref{tbl:app-indicNLG-varta} respectively.\footnote{These tables only contain the results for \dataset/-T5 and the mT5 model we finetuned. The detailed results of the other two models can be found in \citet{indicnlg}.} For paraphrase generation, we report three BLEU variants: standard BLEU, self-BLEU, and iBLEU (with $\alpha=0.7$). We see negative self-BLEU numbers for both models which suggest that, in some languages, there is a large lexical overlap between the input sentences and the generated paraphrases.

\begin{table*}
\small
\centering
\setlength{\tabcolsep}{3pt}
\begin{tabular}{rcccccccccccccrr} 
\toprule
 \multirow{2}{*}{Lang.} & \multicolumn{15}{c}{mT5}\\
 \cmidrule(lr){2-16}
 &\multicolumn{3}{c}{Wikibio} & \multicolumn{3}{c}{Sentence Sum.} & \multicolumn{3}{c}{Head. Gen.} &  \multicolumn{3}{c}{Question Gen.} &  \multicolumn{3}{c}{Paraphrase (BLEU)} \\
  \cmidrule(lr){2-4} \cmidrule(lr){5-7} \cmidrule(lr){8-10} \cmidrule(lr){11-13} \cmidrule(lr){14-16}
 &R1 &  R2 & RL&  R1 &  R2 & RL & R1 &  R2 & RL & R1 &  R2 & RL & BLEU&	SelfBLEU&iBLEU \\
\midrule
as & 57.08 & 37.78 & 55.78 & 38.95 & 23.92 & 37.5 & 34.16 & 18.39 & 32.44 & 22.33 & 8.77 & 21.17 & 20.75 & -12.70 & 10.71 \\
bn & 56.83 & 37.97 & 55.25 & 31.87 & 20.76 & 30.68 & 32.57 & 18.02 & 30.58 & 27.77 & 11.09 & 26.34 & 13.38 & -21.02 & 3.06 \\
gu & - & - & - & 27.3 & 13.6 & 26.13 & 32.96 & 17.49 & 31.39 & 28.59 & 10.91 & 27.46 & 23.71 & -100 & -13.4 \\
hi & 66.68 & 53.06 & 65.79 & 33.52 & 15.92 & 29.81 & 36.61 & 17.61 & 32.16 & 35.57 & 15.71 & 32.73 & 35.13 & -100.0 & -5.40 \\
kn & 42.42 & 22.42 & 40.73 & 73.35 & 67.18 & 72.86 & 47.4 & 30.64 & 46.15 & 24.44 & 9.35 & 23.4 & 17.52 & 0.0 & 12.26 \\
ml & 43.16 & 22.13 & 37.83 & 42.54 & 28.28 & 41.3 & 40.14 & 23.52 & 38.58 & 22.62 & 8.35 & 21.34 & 10.37 & -69.97 & -13.73 \\
mr & - & - & - & 23.33 & 10.4 & 22.61 & 32.59 & 16.89 & 31.48 & 23.78 & 9.03 & 22.92 & 22.99 & -33.43 & 6.06 \\
or & 70.85 & 57.06 & 70.07 & 23.44 & 12.17 & 22.6 & 22.55 & 10.25 & 21.56 & 25.83 & 10.88 & 24.51 & 19.94 & -100.0 & -16.03 \\
pa & 54.78 & 34.95 & 53.33 & 48.78 & 32.16 & 45.32 & 46.85 & 30.06 & 43.4 & 33.19 & 13.73 & 31.13 & 25.32 & -63.89 & -1.44 \\
ta & 52.24 & 30.11 & 50.47 & 40.53 & 22.99 & 38.86 & 47.72 & 31.3 & 46.16 & 23.82 & 8.74 & 22.89 & 18.68 & -100.0 & 16.92 \\
te & 53.17 & 29.56 & 50.65 & 30.44 & 15.45 & 29.43 & 32.22 & 15.77 & 30.83 & 26.72 & 10.64 & 25.56 & 18.14 & -39.76 & 0.76 \\
Avg & 55.25 & 36.12 & 53.32 & 37.64 & 23.89 & 36.1 & 36.89 & 20.9 & 34.98 & 26.79 & 10.65 & 25.4 & 20.54 & -58.25 & -0.02 \\
\bottomrule
\end{tabular}
\caption{Performance of mT5 on IndicNLG tasks.}
\label{tbl:app-indicNLG-mt5}
\end{table*}

\begin{table*}
\small
\centering
\setlength{\tabcolsep}{3pt}
\begin{tabular}{rcccccccccccccrr} 
\toprule
 \multirow{2}{*}{Lang.} & \multicolumn{15}{c}{\dataset/-T5-base-1M}\\
 \cmidrule(lr){2-16}
 &\multicolumn{3}{c}{Wikibio} & \multicolumn{3}{c}{Sentence Sum.} & \multicolumn{3}{c}{Head. Gen.} &  \multicolumn{3}{c}{Question Gen.} &  \multicolumn{3}{c}{Paraphrase (BLEU)} \\
  \cmidrule(lr){2-4} \cmidrule(lr){5-7} \cmidrule(lr){8-10} \cmidrule(lr){11-13} \cmidrule(lr){14-16}
 &R1 &  R2 & RL&  R1 &  R2 & RL & R1 &  R2 & RL & R1 &  R2 & RL & BLEU&	SelfBLEU&iBLEU \\
\midrule
as & 60.56 & 41.4 & 59.1 & 35.67 & 20.05 & 34.03 & 37.61 & 20.8 & 35.54 & 23.58 & 10.19 & 22.48 & 15.17 & -70.71 & -10.59 \\
bn & 58.4 & 39.4 & 56.65 & 31.88 & 17.06 & 29.93 & 39.27 & 20.8 & 36.27 & 35.18 & 15.51 & 32.93 & 16.77 & -22.59 & 4.96 \\
gu & - & - & - & 27.73 & 13.78 & 26.5 & 36.36 & 19.32 & 34.49 & 30.97 & 12.52 & 29.75 & 31.09 & -100.0 & -8.23\\
hi & 67.48 & 54.06 & 66.61 & 34.7 & 16.27 & 30.46 & 41.65 & 21 & 36.37 & 41.01 & 19.56 & 37.42 & 42.51 & -80.91 & 5.48\\
kn & 42.57 & 21.62 & 40.18 & 42.89 & 27.45 & 41.54 & 43.56 & 24.84 & 41.86 & 26.72 & 10.83 & 25.52 & 22.87 & 0.0 & 16.01\\
ml & 44.32 & 23.01 & 38.8 & 37.75 & 23.43 & 36.64 & 39.18 & 21.95 & 37.56 & 24.73 & 9.46 & 23.35 & 19.95 & -100 & -16.04 \\
mr & - & - & - & 23.69 & 10.8 & 22.97 & 36.38 & 19.43 & 35.03 & 26.56 & 10.91 & 25.68 & 31.08 & -50.81 & 6.52  \\
or & 72.78 & 58.8 & 71.73 & 23.62 & 12.35 & 22.7 & 30.13 & 15.04 & 28.76 & 30.77 & 14.37 & 29.23 & 29.03 & -70.71 & -0.89\\
pa & 55.04 & 35.33 & 53.45 & 48.39 & 31.76 & 44.93 & 50.67 & 33.44 & 47.1 & 36.36 & 16.21 & 34.16 & 32.37 & -59.46 & 4.82 \\
ta & 53.26 & 31.49 & 51.41 & 41.06 & 23.35 & 39.36 & 49.35 & 32.44 & 47.7 & 26.04 & 10.08 & 24.98 & 23.97 & -45.18 & 3.23 \\
te & 54.7 & 31.38 & 52.5 & 30.47 & 15.14 & 29.32 & 36.7 & 18.85 & 35.17 & 28.92 & 12.24 & 27.61 & 24.04 & -9.82 & 13.88\\
Avg & 56.57 & 37.39 & 54.49 & 34.35 & 19.22 & 32.58 & 40.08 & 22.54 & 37.8 & 30.08 & 12.9 & 28.46 & 26.26 & -55.47 & 1.74\\
\bottomrule
\end{tabular}
\caption{Performance of \dataset/-T5 on IndicNLG tasks.}
\label{tbl:app-indicNLG-varta}
\end{table*}

\section{Datasheet}\label{sec:datasheet}
We also provide a Datasheet \cite{datasheets} for our dataset in Table 25. This provides an in-depth rationale behind the creation, distribution, and maintenance of \dataset/, including any underlying assumptions, potential risks or harms, and implications of use.

\begin{table*}[t!]
    \center{\footnotesize %
    \begin{tabular}{p{15cm}}
    \toprule 
    \toprule
    \multicolumn{1}{c}{\textbf{Datasheet for \dataset/}} \\
    \midrule
    \textbf{Motivation} \\
    \textbf{Q:} For what purpose was the dataset created? (Was there a specific task in mind? Was there a specific gap that needed to be filled? Please provide a description.) \\
    \textbf{A:} The dataset was created for advancing research on the task of headline generation in Indic languages. There is a need for a large-scale, high-quality dataset in these languages as motivated in \S\ref{sec:intro}. \\  \\
    
    \textbf{Q:} Who created this dataset (e.g., which team, research group) and on behalf of which entity (e.g., company, institution, organization)? \\
    \textbf{A:} No answer as it will violate the anonymity of the authors. \\  \\
    
    \textbf{Q:} Who funded the creation of the dataset? (If there is an associated grant, please provide the name of the grantor and the grant name and number.) \\
    \textbf{A:} No answer as it may violate the anonymity of the authors. \\  \\
     
    \textbf{Q:} Any other comments? \\
    \textbf{A:} None. \\
    \midrule
    \textbf{Composition} \\
    \textbf{Q:} What do the instances that comprise the dataset represent (e.g., documents, photos, people, countries)? (Are there multiple types of instances (e.g., movies, users, and ratings; people and interactions between them; nodes and edges)? Please provide a description.) \\
    \textbf{A:} Each instance contains a news article and its headline in one of 15 languages shown in Table \ref{tbl:app-lang-split}. \\  \\
    
    \textbf{Q:} How many instances are there in total (of each type, if appropriate)? \\
    \textbf{A:} Refer to Table \ref{tbl:app-lang-split} for the number of instances in the train, validation, and test splits of each language. \\  \\
    
    \textbf{Q:} Is any information missing from individual instances? (If so, please provide a description, explaining why this information is missing (e.g., because it was unavailable). This does not include intentionally removed information, but might include, e.g., redacted text.) \\
    \textbf{A:} No \\  \\
    
    \textbf{Q:} Are relationships between individual instances made explicit (e.g., users' movie ratings, social network links)? ( If so, please describe how these relationships are made explicit.) \\
    \textbf{A:} There might be some relationships among the articles. There may be instances of multiple articles within a language, or across languages discussing the same piece of news, but we do not make an effort to explicitly mark them in any way. \\  \\
    
    \textbf{Q:} Are there recommended data splits (e.g., training, development/validation, testing)? (If so, please provide a description of these splits, explaining the rationale behind them.) \\
    \textbf{A:} Yes, we do provide recommended data splits. There are two training sets: \textsc{full} and \textsc{small}. More details can be found in \S\ref{sec:data-splits}. \\  \\
    
    \textbf{Q:} Are there any errors, sources of noise, or redundancies in the dataset? (If so, please provide a description.) \\
    \textbf{A:} The articles themselves are professionally written and should not contain errors. There should also be no redundancies since we deduplicate the data. But there may be errors introduced during the process of crawling and processing the data such as residual HTML content, embedded javascript, headers and footers not related to the article itself, etc. \\  \\
    
    \textbf{Q:} Is the dataset self-contained, or does it link to or otherwise rely on external resources (e.g., websites, tweets, other datasets)? (If it links to or relies on external resources, a) are there guarantees that they will exist, and remain constant, over time; b) are there official archival versions of the complete dataset (i.e., including the external resources as they existed at the time the dataset was created); c) are there any restrictions (e.g., licenses, fees) associated with any of the external resources that might apply to a future user? Please provide descriptions of all external resources and any restrictions associated with them, as well as links or other access points, as appropriate.) \\
    \textbf{A:} The dataset is not self-contained. We only release URLs of the articles due to license restrictions. The dataset can easily be rebuilt using the code we provide. There may be restrictions on sharing the built dataset, but there are none for using them for research purposes. We make sure that the articles will continue to be accessible in the future by providing links to their archived versions on the Wayback Machine. \\  \\
    
    \textbf{Q:} Does the dataset contain data that might be considered confidential (e.g., data that is protected by legal privilege or by doctor-patient confidentiality, data that includes the content of individuals' non-public communications)? (If so, please provide a description.) \\
    \textbf{A:} No; all articles in the dataset are from public news portals. \\
    \end{tabular}     
    }
    \label{tbl:app-datasheet}
    \caption{Datasheet for \dataset/.}
    \end{table*}
    
    \begin{figure*}[t!]
    \center{\footnotesize %
    \begin{tabular}{p{15cm}}
    \textbf{Q:} Does the dataset contain data that, if viewed directly, might be offensive, insulting, threatening, or might otherwise cause anxiety? (If so, please describe why.) \\
    \textbf{A:} Since the dataset contains news articles, there may be reports of violence, murder, racial discrimination, and other triggering phenomena. \\  \\
    
    \textbf{Q:} Does the dataset relate to people? (If not, you may skip the remaining questions in this section.) \\
    \textbf{A:} Yes, most of the articles talk about events involving real people. \\  \\
    
    \textbf{Q:} Does the dataset identify any subpopulations (e.g., by age, gender)? (If so, please describe how these subpopulations are identified and provide a description of their respective distributions within the dataset.) \\
    \textbf{A:} This is not explicitly identified, though many of the articles explicitly mention the gender of the people described/discussed. \\  \\
    
    \textbf{Q:} Is it possible to identify individuals (i.e., one or more natural persons), either directly or indirectly (i.e., in combination with other data) from the dataset? (If so, please describe how.) \\
    \textbf{A:} Yes; their names are given in the running text. \\  \\
    
    \textbf{Q:} Does the dataset contain data that might be considered sensitive in any way (e.g., data that reveals racial or ethnic origins, sexual orientations, religious beliefs, political opinions or union memberships, or locations; financial or health data; biometric or genetic data; forms of government identification, such as social security numbers; criminal history)? (If so, please provide a description.) \\
    \textbf{A:} Yes, some articles may contain the personal information of individuals. But it is very unlikely that they contain information that is not already public. \\  \\
    
    \textbf{Q:} Any other comments? \\
    \textbf{A:} None.\\
    \midrule
    \textbf{Collection Process} \\
    \textbf{Q:} How was the data associated with each instance acquired? (Was the data directly observable (e.g., raw text, movie ratings), reported by subjects (e.g., survey responses), or indirectly inferred/derived from other data (e.g., part-of-speech tags, model-based guesses for age or language)? If data was reported by subjects or indirectly inferred/derived from other data, was the data validated/verified? If so, please describe how.) \\
    \textbf{A:} The data was all downloaded directly from the news aggregator DailyHunt. \\  \\
    
    \textbf{Q:} What mechanisms or procedures were used to collect the data (e.g., hardware apparatus or sensor, manual human curation, software program, software API)? (How were these mechanisms or procedures validated?) \\
    \textbf{A:} We use a Python-based crawling program called Scrapy to collect the data. It is publicly available at this URL: \url{https://scrapy.org/}. The procedure is manually validated by randomly selecting articles and comparing their content with the processed text. \\  \\
    
    \textbf{Q:} If the dataset is a sample from a larger set, what was the sampling strategy (e.g., deterministic, probabilistic with specific sampling probabilities)? \\
    \textbf{A:} The dataset is not part of a larger corpus. \\  \\
    
    \textbf{Q:} Who was involved in the data collection process (e.g., students, crowd workers, contractors) and how were they compensated (e.g., how much were crowd workers paid)? \\
    \textbf{A:} The data is automatically crawled. \\  \\
    
    \textbf{Q:} Over what timeframe was the data collected? (Does this timeframe match the creation timeframe of the data associated with the instances (e.g., recent crawl of old news articles)? If not, please describe the timeframe in which the data associated with the instances was created.) \\
    \textbf{A:} The data was crawled in August 2022. The crawled articles were published between January 2010 to August 2022. \\  \\
    
    \textbf{Q:} Were any ethical review processes conducted (e.g., by an institutional review board)? (If so, please provide a description of these review processes, including the outcomes, as well as a link or other access point to any supporting documentation.) \\
    \textbf{A:} No ethical review processes were conducted. \\  \\
    
    \textbf{Q:} Does the dataset relate to people? (If not, you may skip the remaining questions in this section.) \\
    \textbf{A:} Yes, most of the articles talk about events involving real people. \\  \\
    
    \textbf{Q:} Did you collect the data from the individuals in question directly, or obtain it via third parties or other sources (e.g., websites)? \\
    \textbf{A:} The data was collected from other sources. \\  \\
    
    \textbf{Q:} Were the individuals in question notified about the data collection? (If so, please describe (or show with screenshots or other information) how notice was provided, and provide a link or other access point to, or otherwise reproduce, the exact language of the notification itself.) \\
    \textbf{A:} No, they were not notified. \\
    \end{tabular}     
    }
    \end{figure*}
    
    \begin{figure*}[t!]
    \center{\footnotesize %
    \begin{tabular}{p{15cm}}
    \textbf{Q:} Did the individuals in question consent to the collection and use of their data? (If so, please describe (or show with screenshots or other information) how consent was requested and provided, and provide a link or other access point to, or otherwise reproduce, the exact language to which the individuals consented.) \\
    \textbf{A:} No; since all the news articles are public. \\  \\
    
    \textbf{Q:} If consent was obtained, were the consenting individuals provided with a mechanism to revoke their consent in the future or for certain uses? (If so, please provide a description, as well as a link or other access point to the mechanism (if appropriate).) \\
    \textbf{A:} Not applicable. \\  \\
    
    \textbf{Q:} Has an analysis of the potential impact of the dataset and its use on data subjects (e.g., a data protection impact analysis) been conducted? (If so, please provide a description of this analysis, including the outcomes, as well as a link or other access point to any supporting documentation.) \\
    \textbf{A:} No. \\  \\
    
    \textbf{Q:} Any other comments? \\
    \textbf{A:} None. \\
    \midrule
    \textbf{Preprocessing/cleaning/labeling} \\
    \textbf{Q:} Was any preprocessing/cleaning/labeling of the data done (e.g., discretization or bucketing, tokenization, part-of-speech tagging, SIFT feature extraction, removal of instances, processing of missing values)? (If so, please provide a description. If not, you may skip the remainder of the questions in this section.) \\
    \textbf{A:} Yes; see the `Processing' paragraph in Section \S\ref{sec:dailyhunt}. \\  \\
    
    \textbf{Q:} Was the "raw" data saved in addition to the preprocessed/cleaned/labeled data (e.g., to support unanticipated future uses)? (If so, please provide a link or other access point to the "raw" data.) \\
    \textbf{A:} No, the original raw data is not included in the distribution, due to licensing restrictions. \\  \\
    
    \textbf{Q:} Is the software used to preprocess/clean/label the instances available? (If so, please provide a link or other access point.) \\
    \textbf{A:} Yes, we use Beautiful Soup 4 which is available at this URL: \url{https://www.crummy.com/software/BeautifulSoup/}. \\  \\
    
    \textbf{Q:} Any other comments? \\
    \textbf{A:} None. \\  \\
    \midrule
    \textbf{Uses} \\
    \textbf{Q:} Has the dataset been used for any tasks already? (If so, please provide a description.) \\
    \textbf{A:} The dataset has been used to finetune existing models to perform headline generation on Indic languages. It has also been used to pre-train language models. See \S\ref{sec:models} for more details. \\  \\
    
    \textbf{Q:} Is there a repository that links to any or all papers or systems that use the dataset? (If so, please provide a link or other access point.) \\
    \textbf{A:} No. \\  \\
    
    \textbf{Q:} What (other) tasks could the dataset be used for? \\
    \textbf{A:} The dataset could be used for mining articles that talk about the same piece of news within and across languages to perform interesting analyses such as: (i) how the same news is reported in different languages?, and (ii) how an article evolves as time progresses, and when new information becomes available. \\  \\
    
    \textbf{Q:} Is there anything about the composition of the dataset or the way it was collected and preprocessed/cleaned/labeled that might impact future uses? (For example, is there anything that a future user might need to know to avoid uses that could result in unfair treatment of individuals or groups (e.g., stereotyping, quality of service issues) or other undesirable harms (e.g., financial harms, legal risks) If so, please provide a description. Is there anything a future user could do to mitigate these undesirable harms?) \\
    \textbf{A:} No. \\  \\
    
    \textbf{Q:} Are there tasks for which the dataset should not be used? (If so, please provide a description.) \\
    \textbf{A:} As with any large dataset, \dataset/ can be used in a variety of harmful ways. For example, it can be used to build models that generate hate speech, fake news, etc. \\  \\
    
    \textbf{Q:} Any other comments? \\
    \textbf{A:} None. \\
    \end{tabular}     
    }
    \end{figure*}
    
    \begin{figure*}[t!]
    \center{\footnotesize %
    \begin{tabular}{p{15cm}}
    \midrule
    \textbf{Distribution} \\
    \textbf{Q:} Will the dataset be distributed to third parties outside of the entity (e.g., company, institution, organization) on behalf of which the dataset was created? (If so, please provide a description.) \\
    \textbf{A:} Yes, the dataset will be freely available. \\  \\
    
    \textbf{Q:} How will the dataset be distributed (e.g., tarball on website, API, GitHub)? (Does the dataset have a digital object identifier (DOI)?) \\
    \textbf{A:} The dataset will be made available on GitHub. \\  \\
    
    \textbf{Q:} When will the dataset be distributed? \\
    \textbf{A:} The dataset is distributed after this work is published. \\  \\
    
    \textbf{Q:} Will the dataset be distributed under a copyright or other intellectual property (IP) license, and/or under applicable terms of use (ToU)? (If so, please describe this license and/or ToU, and provide a link or other access point to, or otherwise reproduce, any relevant licensing terms or ToU, as well as any fees associated with these restrictions.) \\
    \textbf{A:} The dataset is licensed under a CC-0 license. \\  \\
    
    \textbf{Q:} Have any third parties imposed IP-based or other restrictions on the data associated with the instances? (If so, please describe these restrictions, and provide a link or other access point to, or otherwise reproduce, any relevant licensing terms, as well as any fees associated with these restrictions.) \\
    \textbf{A:} The news aggregator from which we collect data grants a non-exclusive and non-transferable license to download. More information can be found here: \url{https://www.dailyhunt.com/user-agreement}. \\  \\
    
    \textbf{Q:} Do any export controls or other regulatory restrictions apply to the dataset or individual instances? (If so, please describe these restrictions, and provide a link or other access point to, or otherwise reproduce, any supporting documentation.) \\
    \textbf{A:} Not to our knowledge. \\  \\
    
    \textbf{Q:} Any other comments? \\
    \textbf{A:} None. \\
    \midrule
    \textbf{Maintenance} \\
    \textbf{Q:} Who is supporting/hosting/maintaining the dataset? \\
    \textbf{A:} All authors will maintain the dataset and the first author will host it on GitHub.\\  \\
    
    \textbf{Q:} How can the owner/curator/manager of the dataset be contacted (e.g., email address)? \\
    \textbf{A:} The E-mail address of the first author will be provided.\\  \\
    
    \textbf{Q:} Is there an erratum? (If so, please provide a link or other access point.) \\
    \textbf{A:} No.\\  \\
    
    \textbf{Q:} Will the dataset be updated (e.g., to correct labeling errors, add new instances, delete instances')? (If so, please describe how often, by whom, and how updates will be communicated to users (e.g., mailing list, GitHub)?) \\
    \textbf{A:} Currently, there is no plan to update the dataset.\\  \\
    
    \textbf{Q:} If the dataset relates to people, are there applicable limits on the retention of the data associated with the instances (e.g., were individuals in question told that their data would be retained for a fixed period of time and then deleted)? (If so, please describe these limits and explain how they will be enforced.) \\
    \textbf{A:} No.\\  \\
    
    \textbf{Q:} Will older versions of the dataset continue to be supported/hosted/maintained? (If so, please describe how. If not, please describe how its obsolescence will be communicated to users.) \\
    \textbf{A:} Yes, if a new update is released, the older versions will still be maintained.\\  \\
    
    \textbf{Q:} If others want to extend/augment/build on/contribute to the dataset, is there a mechanism for them to do so? (If so, please provide a description. Will these contributions be validated/verified? If so, please describe how. If not, why not? Is there a process for communicating/distributing these contributions to other users? If so, please provide a description.) \\
    \textbf{A:} Since the data is collected automatically, others can easily extend/augment/build on/contribute to the dataset. It does not require any additional verification. There is currently no official channel for communication between the authors and potential contributors to communicate. One will be created if it becomes necessary.\\  \\
    
    \textbf{Q:} Any other comments? \\
    \textbf{A:} None. \\
    \bottomrule
    \bottomrule
    \end{tabular}     
    }
    \end{figure*}

\end{document}